\documentclass[10pt,journal,compsoc]{IEEEtran}
\ifCLASSOPTIONcompsoc
  \usepackage[nocompress]{cite}
\else
  \usepackage{cite}
\fi
\usepackage{multirow}
\usepackage{booktabs} 
\usepackage{colortbl}
\usepackage{pifont}
\usepackage{adjustbox}	
\usepackage{url}
\usepackage{tabu}
\usepackage{amssymb}
\usepackage{helvet} 
\usepackage{textcomp} 
\usepackage{paralist}

\usepackage[T1]{fontenc}
\usepackage[para,online,flushleft]{threeparttable}

\usepackage{amsmath,empheq}
\usepackage{calrsfs}
\DeclareMathAlphabet{\pazocal}{OMS}{zplm}{m}{n}
\usepackage{amsfonts}
\usepackage{enumitem}
\usepackage{algorithm}
\usepackage{algorithmic}
\usepackage{color}
\usepackage[dvipsnames]{xcolor}
\usepackage{colortbl}
\usepackage[most]{tcolorbox}
\usepackage{tikz}
\usepackage[framemethod=TikZ]{mdframed}
\usepackage{calculator}

\usepackage{array}
\newcolumntype{L}[1]{>{\raggedright\let\newline\\\arraybackslash\hspace{0pt}}m{#1}}
\newcolumntype{C}[1]{>{\centering\let\newline\\\arraybackslash\hspace{0pt}}m{#1}}
\newcolumntype{R}[1]{>{\raggedleft\let\newline\\\arraybackslash\hspace{0pt}}m{#1}}

\ifCLASSINFOpdf
\else
\fi
\begin{document}
\bstctlcite{IEEEexample:BSTcontrol}
%
\title{Learning-From-Disagreement: A Model Comparison and Visual Analytics Framework}
%
%
%
%

\author{Junpeng~Wang,
	Liang~Wang,
	Yan~Zheng,
	Chin-Chia~Michael~Yeh,
  Shubham~Jain,
  and~Wei~Zhang
\IEEEcompsocitemizethanks{\IEEEcompsocthanksitem J. Wang, L. Wang, Y. Zheng, C.-C. M. Yeh, S. Jain, and W. Zhang are with Visa Research, Palo Alto,
CA, 94301.\protect\\
E-mail: \{junpenwa, liawang, yazheng, miyeh, shubhjai, wzhan\}@visa.com}
\thanks{Manuscript received April xx, 20xx; revised August xx, 20xx.}}

%
%

\markboth{Journal of \LaTeX\ Class Files,~Vol.~xx, No.~x, August~20xx}%
{Shell \MakeLowercase{\textit{et al.}}: Bare Demo of IEEEtran.cls for Computer Society Journals}
%



\newcommand{\sysname}{\textit{LFD}ive}

\definecolor{clrstep1}{rgb}{0.925, 0.925, 0.925}
\definecolor{clrstep2}{rgb}{0.902, 0.937, 0.851}
\definecolor{clrstep3}{rgb}{0.957, 0.898, 0.839}
\definecolor{clrstep4}{rgb}{0.984, 0.949, 0.8}
\definecolor{clrstep5}{rgb}{0.878, 0.918, 0.969}
\definecolor{clrstep6}{rgb}{0.922, 0.894, 0.941}

\definecolor{darkgreen}{rgb}{0.0, 0.7, 0.0}
\definecolor{boxblue}{rgb}{0.875, 0.922, 0.973}
\definecolor{boxgreen}{rgb}{0.890, 0.941, 0.843}

\definecolor{boxsingle}{rgb}{0.875, 0.922, 0.973}
\definecolor{boxensemble}{rgb}{0.875, 0.922, 0.973}

\newcommand{\etal}{\textit{et al.}}
\newcommand{\fd}{\textcolor{red}}
\newcommand{\fdb}{\textcolor{blue}}

\newtcbox{\boxsingle}{on line,
  colframe=gray,        
  colback=red!5,
  coltext=black,        
  boxrule=0.1pt,        
  arc=0pt,              
  boxsep=0.1pt,
  left=1pt,right=1pt,top=0.5pt,bottom=0.5pt,
}

\newtcbox{\boxensemble}{on line,
  colframe=gray,        
  colback=green!5,
  coltext=black,        
  boxrule=0.1pt,        
  arc=0pt,              
  boxsep=0.1pt,
  left=1pt,right=1pt,top=0.5pt,bottom=0.5pt,
}

\newtcbox{\boxblue}{on line,
  colframe=gray,        
  colback=boxblue,
  coltext=black,        
  boxrule=0.1pt,        
  arc=0pt,              
  boxsep=0.1pt,
  left=1pt,right=1pt,top=0.5pt,bottom=0.5pt,
}

\newtcbox{\boxgreen}{on line,
  colframe=gray,        
  colback=boxgreen,
  coltext=gray,        
  boxrule=0.1pt,        
  arc=0pt,              
  boxsep=0.1pt,
  left=1pt,right=1pt,top=2pt,bottom=0.5pt,
}

\newtcbox{\rcbox}{on line,
  colframe=white,        
  colback=Cerulean!80!white,
  coltext=white,        
  boxrule=0.1pt,        
  arc=2pt,              
  boxsep=0.1pt,
  left=2pt,right=2pt,top=1pt,bottom=1pt,
}

\newtcbox{\rfbox}{on line,
  colframe=white,        
  colback=Rhodamine!70!white,
  coltext=white,        
  boxrule=0.1pt,        
  arc=2pt,              
  boxsep=0.1pt,
  left=2pt,right=2pt,top=1pt,bottom=1pt,
}

\newtcbox{\ribox}{on line,
  colframe=white,        
  colback=SeaGreen!90!white,
  coltext=white,        
  boxrule=0.1pt,        
  arc=2pt,              
  boxsep=0.1pt,
  left=2pt,right=2pt,top=1pt,bottom=1pt,
}

\newcommand{\steparrow}[4]{
\begin{tikzpicture}[inner sep=0pt,baseline=(base)]
    \DIVIDE{#2}{2}{\halfw}
    \DIVIDE{#3}{2}{\halfh}
    \SUBTRACT{0}{\halfw}{\neghalfw}
    \SUBTRACT{0}{\halfh}{\neghalfh}
    \ADD{\halfw}{\halfh}{\sumhalfwh}
    \ADD{\neghalfw}{\halfh}{\sumneghalfwh}
    \filldraw[color=#1, draw=gray] (\neghalfw,\neghalfh)--(\halfw,\neghalfh)--(\sumhalfwh,0)--(\halfw,\halfh)--(\neghalfw,\halfh)--(\sumneghalfwh,0)--cycle node[text=black] at (0.1, -0.015) {#4};
    \node (base) at (0,-.7ex) {};
\end{tikzpicture}
}

\newcommand{\steparrowthree}[7]{
\begin{tikzpicture}[inner sep=0pt,baseline=(base)]
    \DIVIDE{#2}{2}{\halfw}
    \DIVIDE{#3}{2}{\halfh}
    \SUBTRACT{0}{\halfw}{\neghalfw}
    \SUBTRACT{0}{\halfh}{\neghalfh}
    \ADD{\halfw}{\halfh}{\sumhalfwh}
    \ADD{\neghalfw}{\halfh}{\sumneghalfwh}

    \filldraw[color=#1, draw=gray] (\neghalfw,\neghalfh)--(\halfw,\neghalfh)--(\sumhalfwh,0)--(\halfw,\halfh)--(\neghalfw,\halfh)--(\sumneghalfwh,0)--cycle node[text=black] at (0.1, -0.015) {#4};

    \ADD{#2}{0.1}{\offsetx}
    \ADD{\offsetx}{\neghalfw}{\ptone}
    \ADD{\offsetx}{\neghalfw}{\ptfive}
    \ADD{\offsetx}{\sumneghalfwh}{\ptsix}
    \ADD{\ptone}{0.3}{\pttwo}
    \ADD{\ptsix}{0.3}{\ptthree}
    \ADD{\ptfive}{0.3}{\ptfour}
    \SUBTRACT{\offsetx}{0.3}{\pttext}
    
    \filldraw[color=#5, draw=gray] (\ptone,\neghalfh)--(\pttwo,\neghalfh)--(\ptthree,0)--(\ptfour,\halfh)--(\ptfive,\halfh)--(\ptsix,0)--cycle node[text=black] at (\pttext, -0.015) {#6};
    \node (base) at (0,-.7ex) {};

    \ADD{#2}{0.5}{\offsetxx}
    \ADD{\offsetxx}{\neghalfw}{\ptonex}
    \ADD{\offsetxx}{\neghalfw}{\ptfivex}
    \ADD{\offsetxx}{\sumneghalfwh}{\ptsixx}
    \ADD{\ptonex}{0.3}{\pttwox}
    \ADD{\ptsixx}{0.3}{\ptthreex}
    \ADD{\ptfivex}{0.3}{\ptfourx}
    \SUBTRACT{\offsetxx}{0.3}{\pttextx}
    \ADD{#6}{1}{\textthird}

    \filldraw[color=#7, draw=gray] (\ptonex,\neghalfh)--(\pttwox,\neghalfh)--(\ptthreex,0)--(\ptfourx,\halfh)--(\ptfivex,\halfh)--(\ptsixx,0)--cycle node[text=black] at (\pttextx, -0.015) {\textthird};
    \node (base) at (0,-.7ex) {};
\end{tikzpicture}
}

\newcommand{\steparrowthreegap}[8]{
\begin{tikzpicture}[inner sep=0pt,baseline=(base)]
    \DIVIDE{#2}{2}{\halfw}
    \DIVIDE{#3}{2}{\halfh}
    \SUBTRACT{0}{\halfw}{\neghalfw}
    \SUBTRACT{0}{\halfh}{\neghalfh}
    \ADD{\halfw}{\halfh}{\sumhalfwh}
    \ADD{\neghalfw}{\halfh}{\sumneghalfwh}

    \filldraw[color=#1, draw=gray] (\neghalfw,\neghalfh)--(\halfw,\neghalfh)--(\sumhalfwh,0)--(\halfw,\halfh)--(\neghalfw,\halfh)--(\sumneghalfwh,0)--cycle node[text=black] at (0.1, -0.015) {#4};

    \ADD{#2}{0.1}{\offsetx}
    \ADD{\offsetx}{\neghalfw}{\ptone}
    \ADD{\offsetx}{\neghalfw}{\ptfive}
    \ADD{\offsetx}{\sumneghalfwh}{\ptsix}
    \ADD{\ptone}{0.3}{\pttwo}
    \ADD{\ptsix}{0.3}{\ptthree}
    \ADD{\ptfive}{0.3}{\ptfour}
    \SUBTRACT{\offsetx}{0.3}{\pttext}
    
    \filldraw[color=#5, draw=gray] (\ptone,\neghalfh)--(\pttwo,\neghalfh)--(\ptthree,0)--(\ptfour,\halfh)--(\ptfive,\halfh)--(\ptsix,0)--cycle node[text=black] at (\pttext, -0.015) {#6};
    \node (base) at (0,-.7ex) {};

    \ADD{#2}{0.5}{\offsetxx}
    \ADD{\offsetxx}{\neghalfw}{\ptonex}
    \ADD{\offsetxx}{\neghalfw}{\ptfivex}
    \ADD{\offsetxx}{\sumneghalfwh}{\ptsixx}
    \ADD{\ptonex}{0.3}{\pttwox}
    \ADD{\ptsixx}{0.3}{\ptthreex}
    \ADD{\ptfivex}{0.3}{\ptfourx}
    \SUBTRACT{\offsetxx}{0.3}{\pttextx}

    \filldraw[color=#7, draw=gray] (\ptonex,\neghalfh)--(\pttwox,\neghalfh)--(\ptthreex,0)--(\ptfourx,\halfh)--(\ptfivex,\halfh)--(\ptsixx,0)--cycle node[text=black] at (\pttextx, -0.015) {#8};
    \node (base) at (0,-.7ex) {};
\end{tikzpicture}
}

\newcommand{\steparrowfour}[8]{
\begin{tikzpicture}[inner sep=0pt,baseline=(base)]
    \DIVIDE{#2}{2}{\halfw}
    \DIVIDE{#3}{2}{\halfh}
    \SUBTRACT{0}{\halfw}{\neghalfw}
    \SUBTRACT{0}{\halfh}{\neghalfh}
    \ADD{\halfw}{\halfh}{\sumhalfwh}
    \ADD{\neghalfw}{\halfh}{\sumneghalfwh}

    \filldraw[color=#1, draw=gray] (\neghalfw,\neghalfh)--(\halfw,\neghalfh)--(\sumhalfwh,0)--(\halfw,\halfh)--(\neghalfw,\halfh)--(\sumneghalfwh,0)--cycle node[text=black] at (0.1, -0.015) {#4};

    \ADD{#2}{0.1}{\offsetx}
    \ADD{\offsetx}{\neghalfw}{\ptone}
    \ADD{\offsetx}{\neghalfw}{\ptfive}
    \ADD{\offsetx}{\sumneghalfwh}{\ptsix}
    \ADD{\ptone}{0.3}{\pttwo}
    \ADD{\ptsix}{0.3}{\ptthree}
    \ADD{\ptfive}{0.3}{\ptfour}
    \SUBTRACT{\offsetx}{0.3}{\pttext}
    
    \filldraw[color=#5, draw=gray] (\ptone,\neghalfh)--(\pttwo,\neghalfh)--(\ptthree,0)--(\ptfour,\halfh)--(\ptfive,\halfh)--(\ptsix,0)--cycle node[text=black] at (\pttext, -0.015) {#6};
    \node (base) at (0,-.7ex) {};

    \ADD{#2}{0.5}{\offsetxx}
    \ADD{\offsetxx}{\neghalfw}{\ptonex}
    \ADD{\offsetxx}{\neghalfw}{\ptfivex}
    \ADD{\offsetxx}{\sumneghalfwh}{\ptsixx}
    \ADD{\ptonex}{0.3}{\pttwox}
    \ADD{\ptsixx}{0.3}{\ptthreex}
    \ADD{\ptfivex}{0.3}{\ptfourx}
    \SUBTRACT{\offsetxx}{0.3}{\pttextx}
    \ADD{#6}{1}{\textthird}

    \filldraw[color=#7, draw=gray] (\ptonex,\neghalfh)--(\pttwox,\neghalfh)--(\ptthreex,0)--(\ptfourx,\halfh)--(\ptfivex,\halfh)--(\ptsixx,0)--cycle node[text=black] at (\pttextx, -0.015) {\textthird};
    \node (base) at (0,-.7ex) {};

    \ADD{#2}{0.9}{\offsetxxx}
    \ADD{\offsetxxx}{\neghalfw}{\ptonexx}
    \ADD{\offsetxxx}{\neghalfw}{\ptfivexx}
    \ADD{\offsetxxx}{\sumneghalfwh}{\ptsixxx}
    \ADD{\ptonexx}{0.3}{\pttwoxx}
    \ADD{\ptsixxx}{0.3}{\ptthreexx}
    \ADD{\ptfivexx}{0.3}{\ptfourxx}
    \SUBTRACT{\offsetxxx}{0.3}{\pttextxx}
    \ADD{#6}{2}{\textfourth}

    \filldraw[color=#8, draw=gray] (\ptonexx,\neghalfh)--(\pttwoxx,\neghalfh)--(\ptthreexx,0)--(\ptfourxx,\halfh)--(\ptfivexx,\halfh)--(\ptsixxx,0)--cycle node[text=black] at (\pttextxx, -0.015) {\textfourth};
    \node (base) at (0,-.7ex) {};
\end{tikzpicture}
}




\IEEEtitleabstractindextext{
\begin{abstract}
With the fast-growing number of classification models being produced every day, numerous model interpretation and comparison solutions have also been introduced. For example, LIME~\cite{ribeiro2016should} and SHAP~\cite{NIPS2017_7062} can interpret what input features contribute more to a classifier's output predictions. Different numerical metrics (e.g., accuracy) can be used to easily compare two classifiers. However, few works can interpret the contribution of a data feature to a classifier in comparison with its contribution to another classifier. This \textit{comparative interpretation} can help to disclose the fundamental difference between two classifiers, select classifiers in different feature conditions, and better ensemble two classifiers. To accomplish it, we propose a \textit{learning-from-disagreement (LFD)} framework to visually compare two classification models. Specifically, LFD identifies data instances with disagreed predictions from two compared classifiers and trains a \textit{discriminator} to learn from the disagreed instances. As the two classifiers' training features may not be available, we train the discriminator through a set of \textit{meta-features} proposed based on certain hypotheses of the classifiers to probe their behaviors. Interpreting the trained discriminator with the SHAP values of different meta-features, we provide actionable insights into the compared classifiers. Also, we introduce multiple metrics to profile the importance of meta-features from different perspectives. With these metrics, one can easily identify meta-features with the most complementary behaviors in two classifiers, and use them to better ensemble the classifiers. We focus on binary classification models in the financial services and advertising industry to demonstrate the efficacy of our proposed framework and visualizations.

\end{abstract}

}

\maketitle

\IEEEdisplaynontitleabstractindextext

%
\IEEEpeerreviewmaketitle


\IEEEraisesectionheading{\section{Introduction and Motivation}\label{sec:introduction}}
\IEEEPARstart{C}{lassification}, i.e., predicting the likelihood of given data instances to be different categories, is a fundamental problem in machine learning (ML). Numerous classification models have been proposed for it, including conventional models~\cite{bishop2006pattern}, ensemble learning models~\cite{sagi2018ensemble}, and deep learning models~\cite{lecun2015deep}. The outstanding performance of these classifiers has made them widely adopted in many real-world applications, e.g., spam filtering and click-through rate (CTR) predictions. Often, for the same problem/application, two or multiple models could be developed. Comparing them and identifying the best one to use in a given context become an increasingly more important problem. 

Although many solutions have been proposed to \textit{interpret} features' importance to a single model~\cite{ribeiro2016should, NIPS2017_7062} or \textit{compare} two models' prediction outcomes~\cite{zhang2018manifold, ren2016squares}, few attempts try to tackle both problems simultaneously (\textit{Challenge 1}), i.e., \textit{comparing two models by interpreting which feature in what condition is more important to one model over the other}. Considering the following scenario in spam filtering, where ML practitioners need to choose between two spam classifiers ($\text{\textit{A}}$ and $\text{\textit{B}}$). Using LIME~\cite{ribeiro2016should}, they found the number of URLs in an email (\textit{n\_url}) is an important feature to model $\text{\textit{A}}$. Applying LIME to $\text{\textit{B}}$, \textit{n\_url} is also important. Comparing $\text{\textit{A}}$ and $\text{\textit{B}}$ with aggregated metrics (e.g., accuracy), both show similar overall performance with small differences. Now, a new email with a large \textit{n\_url} comes (target feature: \textit{n\_url}, condition: large), which classifier will behave better? As one can see here, the interpretation of individual models (i.e., LIME's output) does not help to select models in this scenario, as \textit{n\_url} is important to both. The small difference revealed by the aggregated metrics is not sufficient to make any decisions either. We try to address this challenge by \textit{comparatively interpreting} two classifiers based on their behavior discrepancy in different feature conditions, such that one can select between them in the corresponding conditions.

\begin{figure}[tb]
 \centering 
 \includegraphics[width=\columnwidth]{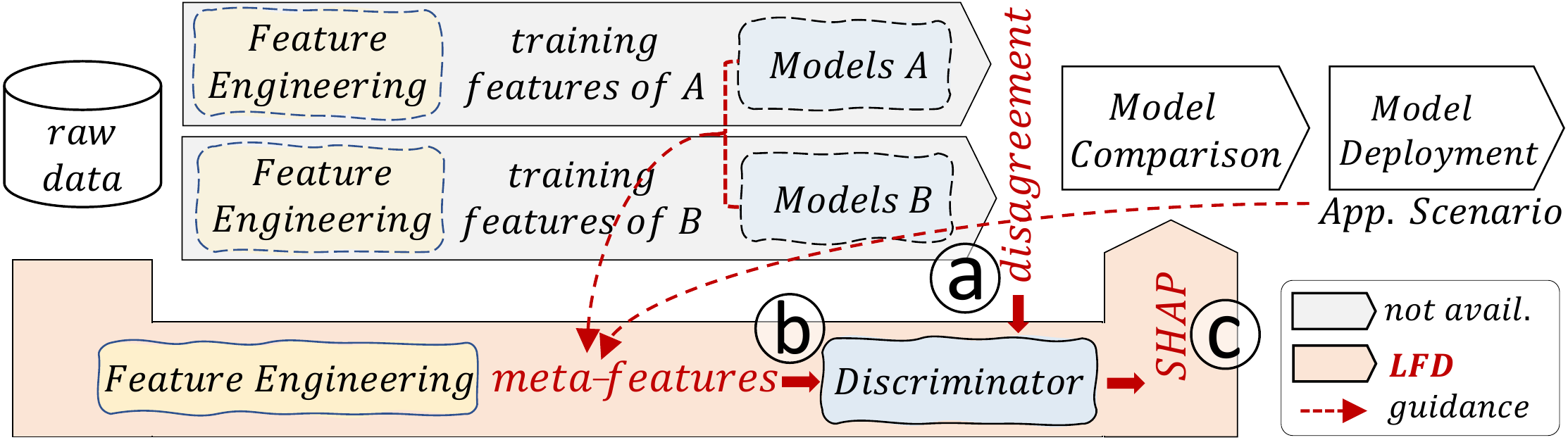}
  \vspace{-0.25in}
 \caption{A general model training, comparison, and deployment pipeline (top). Model-agnostic and feature-agnostic LFD comparisons (bottom).}
 \label{fig:pipeline}
\end{figure}

Another challenge for ML practitioners is that they often do not have full details of the compared models (\textit{Challenge 2}). This is very common in industry, as model training and comparison could be conducted by different teams. Fig.~\ref{fig:pipeline} (top) shows a general model development pipeline, where the training features and architecture details of the two compared models (in gray) are not available. Targeting to serve ML practitioners with sufficient knowledge on their domain data, our work relies on users to propose new features to probe the compared classifiers' behaviors. We call these new features \textit{meta-features}, which can be generated based on: (a) the target application scenario, (b) the general assumptions of the compared models (the two dashed arrows in Fig.~\ref{fig:pipeline}). The spam filtering scenario is a good example for case (a), where users are interested in selecting models based on their behaviors on \textit{n\_url}. They can then derive such a meta-feature, even though \textit{n\_url} may not be used when training either of the two models. For case (b), although users have little details of the compared models, they often have certain hypotheses on the general behaviors of models in the same type. For example, if one of the compared classifiers is an RNN, then sequence-related meta-features (e.g., sequence length) can be proposed to see if the RNN really behaves better on instances with longer sequences. Note that, the meta-features here are generated for interpretations, which differ from the training features (that target to improve models' performance but may be less interpretable). The meta-features could overlap with the training features.

\setlength{\belowcaptionskip}{-5pt}
\setlength{\belowcaptionskip}{0pt}

Combining our ways of tackling the two challenges, we propose \textit{learning-from-disagreement} (LFD) (Fig.~\ref{fig:pipeline}, bottom), a model-agnostic and feature-agnostic framework to compare two classifiers. LFD works in three steps. \textit{First} (Fig.~\ref{fig:pipeline}a), it identifies the prediction disagreement from the two compared classifiers ($\text{\textit{A} and \textit{B}}$) and uses it to label instances. For example, we can give the instances that are captured (highly scored) by $\text{\textit{A}}$ but missed by $\text{\textit{B}}$, denoted as $\text{\textit{A}}^\text{+}\text{\textit{B}}^\text{-}$, as negative. Similarly, $\text{\textit{A}}^\text{-}\text{\textit{B}}^\text{+}$ instances will have positive labels. \textit{Second} (Fig.~\ref{fig:pipeline}b), LFD uses the meta-features of the disagreed instances and their disagreement labels to train a binary classifier, which is named the \textit{discriminator}. This discriminator leverages the power of ML to automatically learn which meta-features contribute more to the disagreement between $\text{\textit{A}}$ and $\text{\textit{B}}$. \textit{Third} (Fig.~\ref{fig:pipeline}c), we use SHAP~\cite{NIPS2017_7062} to conduct feature-based interpretations on the trained discriminator. Our improved SHAP visualizations and scalable meta-feature explorations in this step effectively disclose which model behaves better in what feature conditions.

LFD helps to select models in given feature conditions and directly leads to a better way to combine them. In the spam filtering case, if model $\text{\textit{A}}$ outperforms $\text{\textit{B}}$ when \textit{n\_url} is larger and vice versa, then we can take scores from $A$ for emails with large \textit{n\_url} and scores from $B$ for emails with small \textit{n\_url} to generate a superior ensemble model. Feature-weighted linear stacking (FWLS)~\cite{sill2009feature}, an advanced model ensembling approach, is based on this idea. It dynamically weights up models that behave better in the current instances' feature conditions and weights down the rest. However, when the number of meta-features is large, FWLS becomes hard to train. The meta-features with more complementary behaviors in the two models would ensemble them better, thus should be considered more. To fill the current gap in comprehensive feature-ordering, this work also proposes multiple metrics to rank the meta-features.

To demonstrate the power of LFD and the efficacy of our feature-ordering metrics, we conduct comprehensive case studies on two real-world problems with front-line ML practitioners, (1) merchant category verifications and (2) CTR predictions. Numerous ML classifiers are involved in the comparisons. The insights gained from these studies, the quantitative results, and the feedback from the domain users together validate the efficacy of our approach.

In summary, the contributions of our work include:
\begin{enumerate} 
	\item We propose LFD and facilitate it with visual feature analysis to comparatively interpret a pair of classifiers.

	\item We introduce multiple metrics to prioritize/rank a large number of meta-features from different perspectives.

	\item We present multiple cases of using LFD and the corresponding visualizations to better compare and ensemble a pair of classifiers from real-world applications.
\end{enumerate}

\section{Related Works}
\label{sec:relatedwork}
\textit{\textbf{Interpreting classification models}} is attracting increasingly more attention in recent years and numerous solutions have been proposed~\cite{yuan2021survey, liu2017towards, hohman2018visual, molnar2020interpretable}. Roughly, model interpretations can be categorized into model-specific and model-agnostic. Model-specific interpretations consider classification models as ``white-boxes'', where people have access to all internal details. For example, most interpretations for deep learning models~\cite{strobelt2017lstmvis, wang2018ganviz, kahng2017cti, pezzotti2017deepeyes, liu2016towards, zhou2016learning, springenberg2014striving, wang2018dqnviz, cao2020analyzing, ming2017understanding} visualize and investigate the internal neurons' activation to disclose how data were transformed internally. Model-agnostic interpretations regard predictive models as ``black-boxes'', where only the models' input and output are available~\cite{hohman2019gamut, wexler2019if, krause2016interacting, zhang2018manifold, amershi2015modeltracker, ren2016squares}. These approaches often employ an interpretable surrogate model to mimic or probe the behavior of the interpreted models locally or globally. For example, LIME~\cite{ribeiro2016should} uses a linear model as a surrogate to simulate the local behavior of the complicated classifier to be interpreted. DeepVID~\cite{wang2019deepvid} trains an interpretable model using the knowledge distilled from the original classifier for interpretation. RuleMatrix~\cite{ming2018rulematrix} converts classification models as a set of standardized IF-THEN-ELSE rules using only the models' input-output behaviors. For most of these interpretation solutions, an important goal is to answer \textit{what input features are more important to the models' output}. There are also solutions that statistically quantify features' importance. One outstanding example is the SHapley Additive exPlanation (SHAP)~\cite{NIPS2017_7062,lundberg2018consistent}, which we will explain with details in Sec.~\ref{sec:shap}. Our work is also model-agnostic (and even feature-agnostic), i.e., we assume no knowledge of the compared models or their training features. Moreover, we interpret a model's behavior in contrast to another model, which is more useful when selecting between them.

\textbf{\textit{Comparing classifiers}} is often conducted through numerical metrics, e.g., accuracy and LogLoss. Multiple general model-building and visualization toolkits, e.g., TensorFlow~\cite{abadi2016tensorflow} and scikit-learn~\cite{pedregosa2011scikit}, provide built-in APIs for these metrics. However, these aggregated metrics often fall short to provide sufficient details in model comparison and selection, e.g., two models could achieve the same accuracy in very different ways and the underlying details are often of more interest when comparing them. Many visual analytics works~\cite{piringer2010hypermoval, zhang2018manifold, murugesan2019deepcompare, zeng2017cnncomparator} have tried to go beyond these aggregated metrics for more comprehensive model comparisons. For example, Manifold~\cite{zhang2018manifold} compares two models by disclosing the agreed and disagreed predictions. The comparison is model-agnostic, and for user-selected instances, Manifold can identify the features contributing to their prediction discrepancy.
DeepCompare\cite{murugesan2019deepcompare} compares deep learning models with incomparable architectures (i.e., CNN v.s. RNN) through their activation patterns. CNNComparator~\cite{zeng2017cnncomparator} compares the same CNN from different training stages to reveal the model's evolution. Deconvolution techniques have also been adopted to compare CNNs~\cite{yu2016visualizing}.
These existing comparison works mostly rely on humans' visual comprehension to identify models' behavior differences. In contrast, our LFD framework is a \textit{learning} approach, which leverages the power of ML to compare ML models.

\textbf{\textit{Feature visualization}} in ML focuses either on (1) revealing what features have been captured by predictive models, or (2) prioritizing features based on their importance to limit the scope of analysis. The former is often conducted on image data and a typical example is from Olah \etal~\cite{olah2017feature}, where the authors used ``visualization by optimization'' to produce feature maps that activate different neurons to interpret deep learning models. Different saliency-map generation algorithms~\cite{zhou2016learning,montavon2017explaining,selvaraju2017grad} also share the same goal of highlighting the captured features to better understand deep neural networks. 
The latter focus of feature prioritization is often conducted on tabular data, where different metrics are used to order the contributions of different data features~\cite{zhang2018manifold, liu2017visual, lundberg2018consistent, muhlbacher2013partition}. For example, when interpreting tree-based models, the number of times that each feature was used to split a tree node is often used to rank the features~\cite{loh2011classification, chen2016xgboost, wang2020investigating}. LIME~\cite{ribeiro2016should} and SHAP~\cite{NIPS2017_7062,lundberg2018consistent} also provide quantitative metrics to order different data features. As the data of our focused applications are in tabular format, we study more on the feature ordering part in this work, and we contribute several new metrics (Sec.~\ref{sec:featureorder}) to prioritize features based on their complementary-level in two compared classifiers.

\section{Background and Motivation}
Our objective is to (1) compare two classifiers; (2) reveal how different features behave in them; and (3) use the insights from (1) and (2) to better ensemble the models. This section explains the necessary background for these three topics.

\subsection{Comparing Classification Models}
Comparing two classifiers is a very common and frequent operation in industry. For example, replacing an old production model with a new model often comes with significant business impacts. Therefore, thorough studies need to be conducted beforehand to compare the two models. These studies are often referred to as ``swap analysis'', deciding whether the old model should be replaced/swapped by the new one. We focus on this application scenario in this work and thus target on the comparison between two models.

To date, most of the model comparisons are limited to numerical metrics, e.g., the area under the receiver operating characteristic (ROC) curve (AUC) and LogLoss. However, these metrics cannot tell in what feature conditions one model outperforms the other, which is a critical question for model selections (e.g., the spam filtering scenario in Sec.~\ref{sec:introduction}). Furthermore, as very limited information is available during comparisons, model-agnostic and feature-agnostic comparisons are often more preferable.

\subsection{Feature Analysis with SHAP}
\label{sec:shap}
Feature analysis is to reveal how an input feature impacts the output of a classifier and the magnitude of the impact. We use one of the most popular solutions, i.e., SHAP, for this analysis, as it is \textit{consistent} and \textit{additive}~\cite{NIPS2017_7062,lundberg2018consistent,wang2020investigating}.

\begin{figure}[tb]
 \centering 
 \includegraphics[width=0.96\columnwidth]{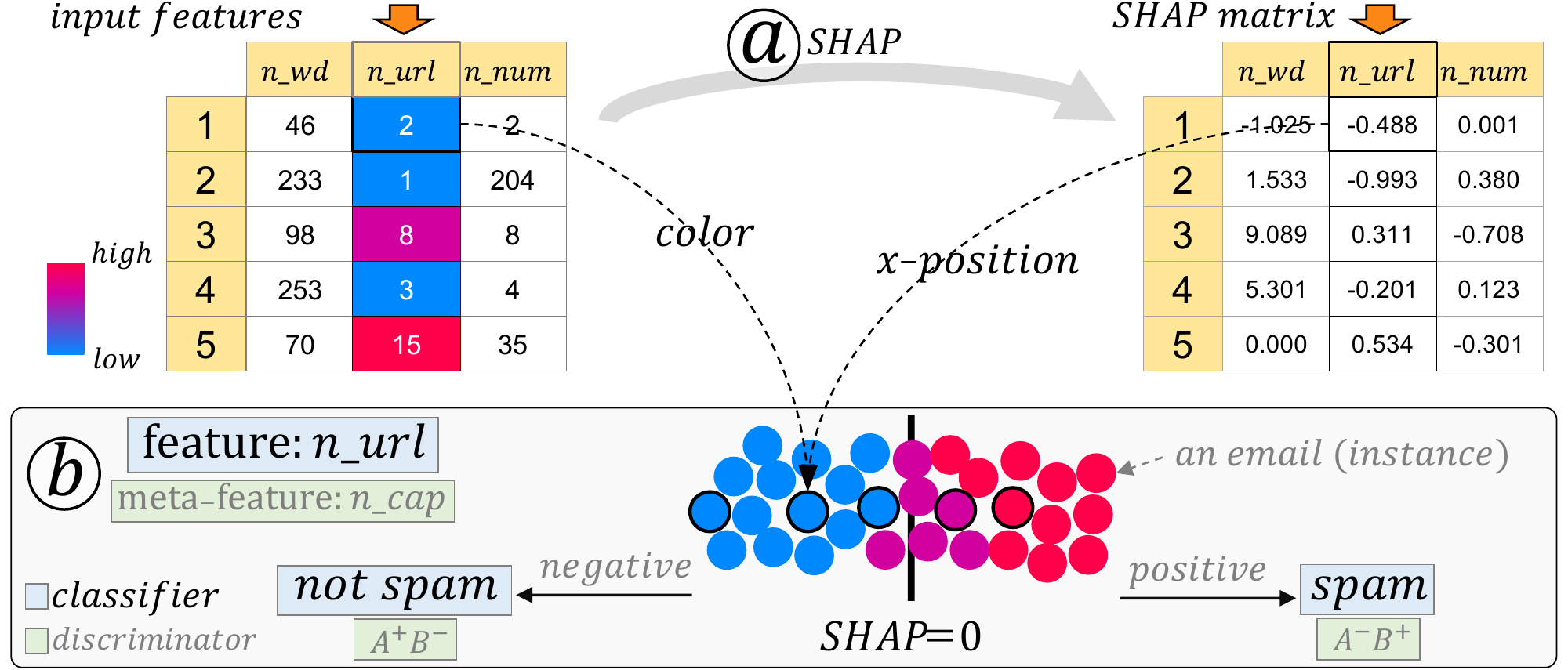}
  \vspace{-0.12in}
 \caption{(a) Interpreting a classifier's behavior on a dataset with SHAP will generate an interpretation matrix sharing the same size with the input. (b) Visualizing the contribution of a feature using the summary-plot~\cite{lundberg2018consistent} (read labels in the \boxblue{blue} boxes). The collective behavior of all instances reflects that emails with higher \textit{n\_url} are more likely to be spam. The summary-plot can also interpret the discriminator from LFD (read labels in the \boxgreen{green} boxes), i.e., emails with higher \textit{n\_cap} (a meta-feature) are more likely to be captured by model $\text{\textit{B}}$ but missed by $\text{\textit{A}}$ (i.e., $\text{\textit{A}}^\text{-}\text{\textit{B}}^\text{+}$).
 }
 \label{fig:shap}
\end{figure}

A tabular dataset with $n$ instances and $m$ features can be considered as a matrix of size $n{\times}m$. 
Fig.~\ref{fig:shap}a (left) presents a fake dataset for a spam classifier. The dataset has 5 instances (emails), each with 3 features: the word count ($n\_wd$), the number of URLs ($n\_url$), and the count of numerical values ($n\_num$) in an email. 
To interpret the classifier's behavior on this dataset, SHAP generates an interpretation matrix, which also has the size of $n{\times}m$ (Fig.~\ref{fig:shap}a, right). Each element of this matrix, $[i, j]$, denotes the contribution of the $j$th feature to the prediction of the $i$th instance. In other words, the sum of all values from the $i$th row of the SHAP matrix is the classifier's final prediction for the $i$th instance  (i.e., the $log(odds)$, which will go through a $sigmoid()$ to become the final probability value). The above explanation considers SHAP as a black-box, as the detail of generating the SHAP matrix is out of the scope of this paper.

The SHAP summary-plot~\cite{lundberg2018consistent} (Fig.~\ref{fig:shap}b, read labels in the \boxblue{blue} boxes) is designed to visualize the effect of individual features to a classifier's prediction. For example, to show the impact of $n\_url$ to the spam classifier, the summary-plot encodes each email as a point. The color and horizontal position of the point reflect the corresponding feature- and SHAP-value respectively. The five points with black stroke in Fig.~\ref{fig:shap}b show the five emails in Fig.~\ref{fig:shap}a. Extending the plot to more emails/instances, we can conclude the feature's impact from the instances' collective behaviors, i.e., emails with more URLs (the red points with larger positive SHAP) are more likely to be spam. The impact of other features can be visualized and vertically aligned with this feature (along the $SHAP{=}0$ line, see Fig.~\ref{fig:order}). These features will be ordered based on their importance to the classifier, which is computed by $mean(|SHAP|){=}\frac{1}{n}\sum_{i=1}^n|SHAP_i|$.

\setlength{\belowcaptionskip}{-12pt}
\setlength{\belowcaptionskip}{0pt}

\begin{figure*}[tbh]
 \centering 
 \includegraphics[width=\textwidth]{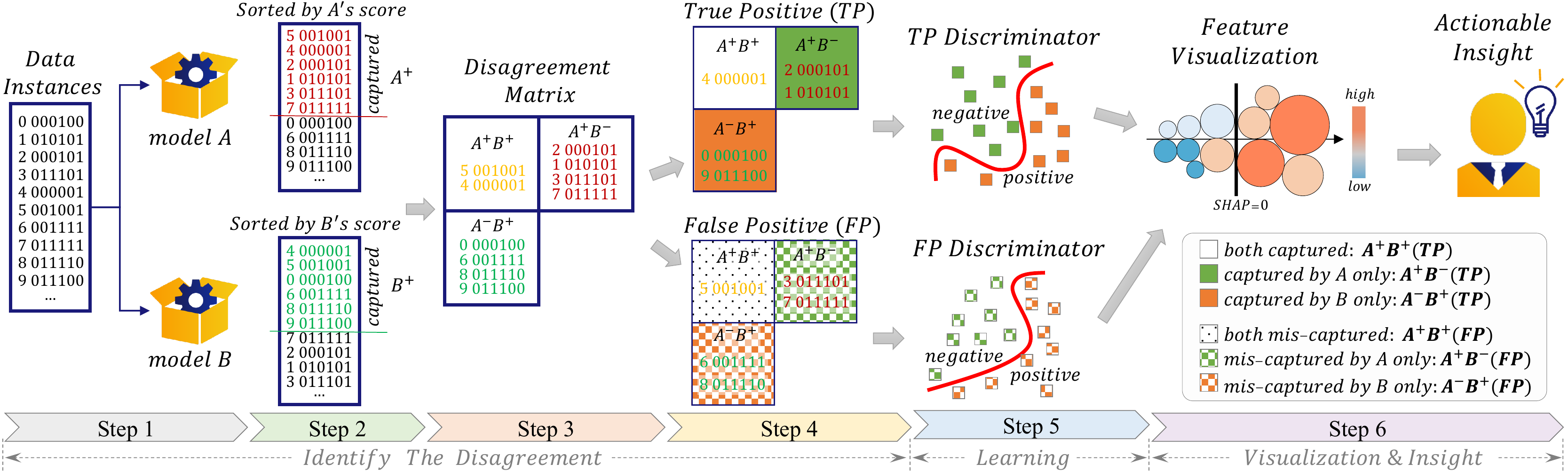}
  \vspace{-0.25in}
 \caption{LFD compares two classifiers ($\text{\textit{A}}$ \& $\text{\textit{B}}$, \textit{Step 1}) by extracting data instances captured by each classifier ($\text{\textit{A}}^\textit{+}$ and $\text{\textit{B}}^\text{+}$, \textit{Step 2}), and joining them into both-captured ($\text{\textit{A}}^\text{+}\text{\textit{B}}^\text{+}$) and captured-only instances ($\text{\textit{A}}^\text{+}\text{\textit{B}}^\text{-}$ and $\text{\textit{A}}^\text{-}\text{\textit{B}}^\text{+}$, \textit{Step 3}). Based on the data labels, the three sets of instances can further be divided into true-positive and false-positive groups (\textit{Step 4}, see the six cells' legend at the bottom-right corner). For each group, a discriminator (binary classifier) is trained, considering $\text{\textit{A}}^\text{-}\textit{\text{B}}^\text{+}$ and $\text{\textit{A}}^\text{+}\text{\textit{B}}^\text{-}$ as positive and negative samples, and using a set of meta-features (\textit{Step 5}). Interpreting the discriminator through SHAP (\textit{Step 6}), we derive insights to effectively interpret which classifier prefers which features in what value ranges.}
 \label{fig:framework}
\end{figure*}

\subsection{Model Ensembling}
\label{sec:bgensemble}
Model ensembling~\cite{zhou2012ensemble} is the research of combining multiple pre-trained models to achieve better performance than individuals. The na\"ive way of ensembling two models is to train a linear regressor to fit scores from the two pre-trained models, which is known as linear stacking~\cite{breiman1996stacked}. Considering the data instance $x$ and two pre-trained models $M_a$ and $M_b$, the linear stacking result $LS(x)$ is computed by:
\begin{equation}\footnotesize
\label{eq:lr}
LS(x) = w_1{\cdot}M_{a}(x) + w_2{\cdot}M_{b}(x).
\end{equation}

Feature-weighted linear stacking (FWLS)~\cite{sill2009feature}, a state-of-the-art model ensembling solution, claims that the weights ($w_1$ and $w_2$ in Eq.~\ref{eq:lr}) should not be fixed but vary based on different feature-values. Because the two models may have varying performances in different feature value ranges. FWLS (Eq.~\ref{eq:fwls}), therefore, combines data features with models' scores by feature-crossing and trains the regressor on the crossed features. The weights $w_1$ and $w_2$ become $m$ pairs of weights ($w_1^i$ and $w_2^i$ in Eq.~\ref{eq:fwls}), where $m$ is the number of features and $F_i(x)$ retrieves the $i$th feature value of $x$.
\begin{equation}\scriptsize
\label{eq:fwls}
FWLS(x) =\sum_{i=1}^{m} [w_{1}^{i}{\cdot}(M_{a}(x){\cdot}F_i(x)) +  w_{2}^{i}{\cdot}(M_{b}(x){\cdot}F_i(x))].
\end{equation}
As the number of features may be large, FWLS can be conducted using a subset of the features. However, incautiously selected features may impair the ensembling result. Therefore, properly ranking the features and choosing the most important ones for ensembling become a critical problem. We propose different metrics to rank features in Sec.~\ref{sec:featureorder}.

\section{Learning-From-Disagreement (LFD)}
\label{sec:lfd}
This section concretizes LFD into six execution steps (Fig.~\ref{fig:framework}). \textit{Step 1$\sim$4} identify the disagreement (Fig.~\ref{fig:pipeline}a). \textit{Step 5} trains a discriminator to learn from the disagreement based on user-proposed meta-features (Fig.~\ref{fig:pipeline}b). \textit{Step 6} interprets the discriminator to answer which model behaves better in what feature conditions (Fig.~\ref{fig:pipeline}c). In detail, the six steps are:

\begin{enumerate}
\item Feed data into the compared classifiers ($\text{\textit{A}}$ \& $\text{\textit{B}}$) and get the two classifiers' scores for individual data instances.

\item Sort instances by the two sets of scores decreasingly, and set a \textit{threshold} as the score cutoff (e.g., 5\% of all instances). Instances with scores above it are those captured by individual models ($\text{\textit{A}}^\text{+}$ and $\text{\textit{B}}^\text{+}$). The threshold often depends on applications, e.g., for loan eligibility predictions, it will be decided by a bank's budget.

\item Join the two sets of captured instances (i.e., $\text{\textit{A}}^\text{+}$ and $\text{\textit{B}}^\text{+}$) into three cells of the \textit{disagreement matrix}, i.e., $\text{\textit{A}}$ captured $\text{\textit{B}}$ missed ($\text{\textit{A}}^\text{+}\text{\textit{B}}^\text{-}$), $\text{\textit{A}}$ missed $\text{\textit{B}}$ captured ($\text{\textit{A}}^\text{-}\text{\textit{B}}^\text{+}$), and both captured ($\text{\textit{A}}^\text{+}\text{\textit{B}}^\text{+}$). For comparison purposes, the $\text{\textit{A}}^\text{+}\text{\textit{B}}^\text{+}$ instances are of less interest. Also, we do not have $\text{\textit{A}}^\text{-}\text{\textit{B}}^\text{-}$ instances, since the filtered instances from \textit{Step 2} are at least captured by one model.

\item Based on the true label of the captured instances, we divide the disagreement matrix into two matrices: one for the TP instances (i.e., correctly captured), the other for the FP instances (i.e., mis-captured).

\item Train the TP and FP discriminators (two binary classifiers) to differentiate the $\text{\textit{A}}^\text{+}\text{\textit{B}}^\text{-}$ (negative) and $\text{\textit{A}}^\text{-}\text{\textit{B}}^\text{+}$ (positive) instances from the TP and FP sides, respectively. The training uses a set of \textit{meta-features}, as the features of model $\text{\textit{A}}$ and $\text{\textit{B}}$ may not be available. 

\item Interpret the trained discriminators with SHAP to provide insights into the fundamental difference between $\text{\textit{A}}$ and $\text{\textit{B}}$. The insights also help to rank the meta-features and pick the best ones to ensemble $\text{\textit{A}}$ and $\text{\textit{B}}$.

\end{enumerate}

\textbf{\textit{Meta-features}} came into the picture, as the features used to train classifier $\text{\textit{A}}$ and $\text{\textit{B}}$ may not be available during comparison (i.e., our second motivating challenge in Sec.~\ref{sec:introduction}). Nevertheless, if the training features are readily available, they can directly work as meta-features. For example, in Fig.~\ref{fig:shap}a, $n\_wd$, $n\_url$, and $n\_num$ are three features derived from raw emails (email title, address, and body) for model training. They can also work as meta-features if users are interested to know how the compared models will behave on them (e.g., which model performs better when $n\_url$ is large?). In cases where the training features are not available, users can derive meta-features based on their interested application scenario. For example, in Fig.~\ref{fig:shap}b, $n\_cap$ (the number of capitalized words) is proposed to \textit{probe} how the compared spam classifiers would be impacted by this meta-feature (though this feature was not used to train either of the classifiers). Meta-features can also be derived based on the type of the compared models. For example, when comparing RNNs with tree-based models, one can generate sequence-related meta-features to verify if the RNNs really benefit from being aware of sequential behaviors. To compare GNNs with RNNs, one can propose neighbor-related meta-features (e.g., nodes' degree) to reveal how much the GNNs can take advantage of neighbors' information. As LFD targets to serve users with sufficient knowledge on their domain data, generating a sufficient number of meta-features to train the discriminator is usually not a bottleneck.

The \textbf{\textit{discriminator}} can be any binary classifier, as long as it is SHAP-friendly (this work uses XGBoost~\cite{chen2016xgboost}). Based on the SHAP of different meta-features, we provide insights into the two compared classifiers. For example, if the discriminator shows the SHAP view in Fig.~\ref{fig:shap}b (use labels in the \boxgreen{green} boxes), we can conclude that ``compared to $\text{\textit{A}}$, classifier $\text{\textit{B}}$ tends to capture emails containing more capitalized words ($n\_cap$ is a meta-feature)''. If the discriminator is the TP discriminator, we know that $\text{\textit{B}}$ outperforms $\text{\textit{A}}$ when this feature is large (correctly captures more). However, if the discriminator is the FP discriminator, $\text{\textit{B}}$ is not as good as $\text{\textit{A}}$, as it mis-captures more when the feature value is large.

LFD has the following direct advantages:
\begin{compactitem}
	\item \textit{\textbf{Model-Agnostic}}: LFD needs only the input and output from the two compared models, making it a generally applicable solution to compare any types of classifiers.

	\item \textit{\textbf{Feature-Agnostic}}: LFD compares classifiers based on newly proposed meta-features, which are independent of the original model-training features. 

	\item \textit{\textbf{Avoiding Data-Imbalance}}: For many real-world applications (e.g., anomaly detections and CTR predictions), data-imbalance, i.e., positive instances are much fewer than negative instances, brings a big challenge to model training and interpretation. LFD smartly avoids this, as it compares the \textit{difference} between two models, i.e., the two ``captured-only'' cells usually have similar sizes.
\end{compactitem}

\section{Domain Experts and Design Requirements}
\label{sec:requirement}
We worked with eight ML experts, specialized in various classification models, to study the model comparison problem. All are ML researchers and front-line model designers in an industrial research lab. Among them, five are research scientists with model training and feature engineering experience for 5+ years. The other three are senior researchers with decades of experience on predictive models. Two experts ($E_1$, $E_2$) intensively participated in the design of LFD, as the framework was originated from their desire to help the ``swap analysis'' between a to-be-launched model and a production model. Two experts ($E_3$, $E_4$) helped to prepare the transaction/CTR data and the ML models we used in Sec.~\ref{sec:casestudy}, but did not use our LFD system until we conduct case studies with them. These four experts co-authored this work. The rest four experts ($E_5{\sim}E_8$) participated in the usability studies only to provide objective feedback.

While working with $E_1$ and $E_2$, we identified multiple requirements for LFD. For example, they mentioned that knowing the distribution of the disagreed instances is where they want to get started. When building different models and analyzing features' contributions, accurate feature importance order is crucial. 
Through intensive discussions in multiple iterations, we distilled the following requirements focused on \textit{model comparison} (\rcbox{MC}) and \textit{feature analysis} (\rfbox{FA}): 

\begin{compactitem}
\item \rcbox{MC1} Present the distribution of data instances across the six cells of the two disagreement matrices (\textit{Step 4} of LFD) under different thresholds (\textit{Step 2} of LFD).
\item \rcbox{MC2} Interpret the TP and FP discriminators to explain the impact of different meta-features on the captured and mis-captured instances (\textit{Step 6} of LFD).
\item \rcbox{MC3} Be able to compare the contribution of the same meta-feature to the TP and FP discriminators, and interpret the difference with as little as possible information.
\item \rfbox{FA1} Disclose the behavior difference of the two compared classifiers on a user-proposed meta-feature.
\item \rfbox{FA2} Effectively rank meta-features from different perspectives and use the more important ones (e.g., more complementary ones) to improve model ensembling.
\end{compactitem}

\setlength{\belowcaptionskip}{-5pt}
\begin{figure*}[tbh]
 \centering 
 \includegraphics[width=\textwidth]{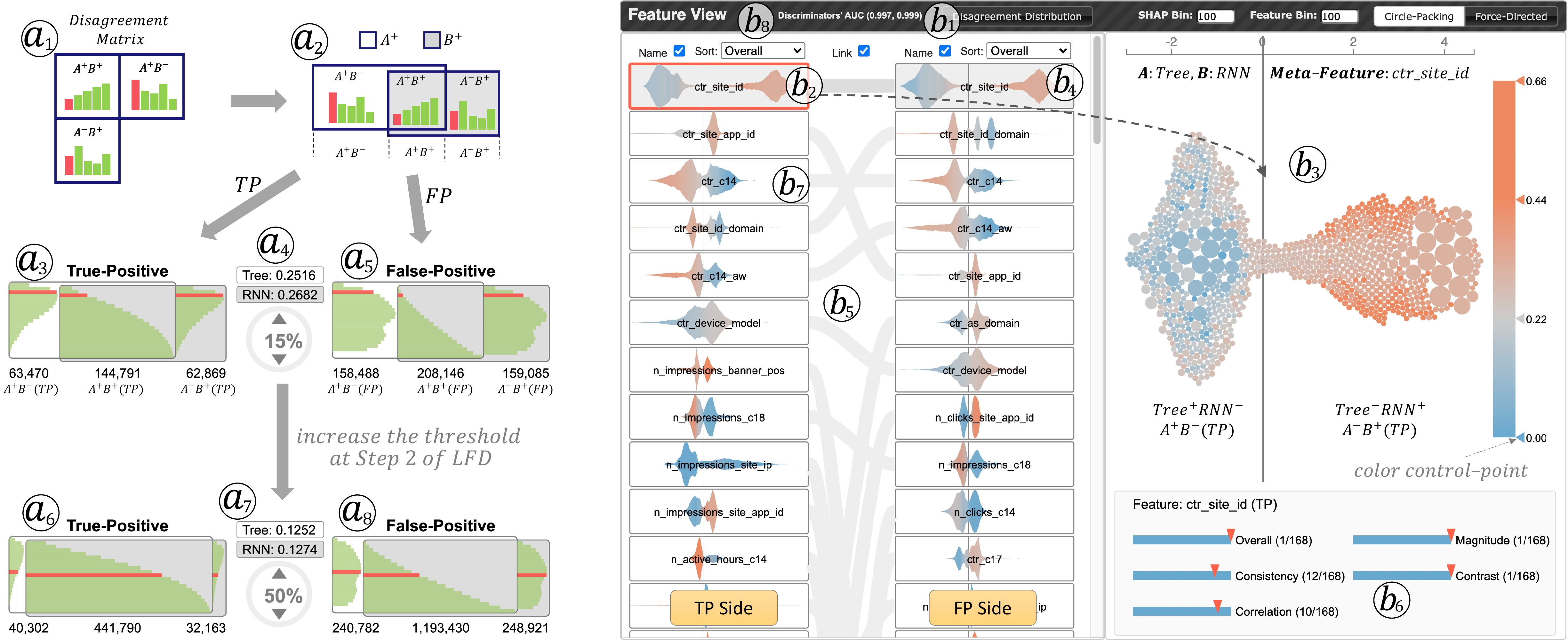}
  \vspace{-0.28in}
 \caption{(a1-a8) The \textit{Disagreement Distribution View} presents the size of the six cells (i.e., $\text{\textit{A}}^\text{+}\text{\textit{B}}^\text{+}$, $\text{\textit{A}}^\text{+}\text{\textit{B}}^\text{-}$, and $\text{\textit{A}}^\text{-}\text{\textit{B}}^\text{+}$ from both the TP and FP sides, \textit{Step 4} of LFD) across different thresholds. The \textit{Feature View} presents the important meta-features to the TP and FP discriminators as two feature lists (b5). Users can select a meta-feature to check its details (b2, b3). (b6) shows the ranks of the selected meta-feature in different metrics.}
 \label{fig:system}
\end{figure*}
\setlength{\belowcaptionskip}{0pt}

\section{Visual Interface}
\label{sec:system}

Following the design requirements, two parts of LFD need to be visually presented: (1) the disagreement matrices under different cutoffs (\textit{Step 2$\sim$4}) and (2) the meta-features and their SHAP values (\textit{Step 6}). These two parts require humans' interactions and visual interpretations on the LFD outcomes. We present them with the \textit{Disagreement Distribution View} and the \textit{Feature View} respectively, in two subsections.

\subsection{Disagreement Distribution View} 
The \textit{Disagreement Distribution View} (Fig.~\ref{fig:system}, left) is designed to meet \rcbox{MC1}. Given a threshold at \textit{Step 2} of LFD, we filter out two sets of instances captured by model $\text{\textit{A}}$ and $\text{\textit{B}}$ respectively. These instances are then joined into three cells at \textit{Step 3}, i.e., $\text{\textit{A}}^\text{+}\text{\textit{B}}^\text{+}$, $\text{\textit{A}}^\text{+}\text{\textit{B}}^\text{-}$, and $\text{\textit{A}}^\text{-}\text{\textit{B}}^\text{+}$. The height of red bars in Fig.~\ref{fig:system}-a1 denotes the size of the three cells under the current threshold. When changing the threshold, the size of the three cells will change accordingly. Thus, we show a sequence of green bars inside each cell, denoting the cell sizes across all possible thresholds (i.e., the horizontal axis of each bar chart represents different thresholds and the vertical axis denotes the number of instances). This distribution overview guides users to select a proper threshold to maximize the size of disagreed data for learning.

To better encode the data joining process, we use a white and a gray background rectangle to represent the instances captured by model $\text{\textit{A}}$ and $\text{\textit{B}}$ respectively, i.e., $\text{\textit{A}}^\text{+}$ and $\text{\textit{B}}^\text{+}$ (Fig.~\ref{fig:system}-a2). Their overlapped region reflects the instances captured by both ($\text{\textit{A}}^\text{+}\text{\textit{B}}^\text{+}$) and the non-overlapped regions on the two sides are the instances captured by one only ($\text{\textit{A}}^\text{+}\text{\textit{B}}^\text{-}$ and $\text{\textit{A}}^\text{-}\text{\textit{B}}^\text{+}$). The width of the rectangles (and the overlapped region) is proportional to the number of instances.

Coming to \textit{Step 4} of LFD, the disagreement matrix in Fig.~\ref{fig:system}-a2 is divided into two matrices when considering the label of the instances. Meanwhile, we rotate the three green bar-charts 90 degrees and present the current threshold value in the middle of the two disagreement matrices. In Fig.~\ref{fig:system}-a4, the threshold is 15\% and the corresponding bars in the six disagreement matrix cells are highlighted in red, Fig.~\ref{fig:system}-a3 (TP side), Fig.~\ref{fig:system}-a5 (FP side).

The two triangles in Fig.~\ref{fig:system}-a4 help to flexibly adjust the threshold (\textit{Step 2} of LFD). The current threshold is 15\% and the cutoff scores for the two compared models, \textit{Tree} ($\text{\textit{A}}$) and \textit{RNN} ($\text{\textit{B}}$), are 0.2516 and 0.2682 respectively. Instances with respective scores larger than these are captured by individual models. When increasing the threshold to 50\% (Fig.~\ref{fig:system}-a7), the overlapped cell becomes larger but the two ``captured-only'' cells become smaller (Fig.~\ref{fig:system}-a6,~\ref{fig:system}-a8). The white and gray rectangles will be overlapped completely, when the threshold becomes 100\% (i.e., both models consider all instances as captured). Note that, the \textit{Disagreement Distribution View} is hidden from the interface by default, but can be enabled by clicking the button in Fig.~\ref{fig:system}-b1.

We have also explored several alternative designs to present the disagreement matrices. For example, we tried to lay out the three cells in the way they are placed in Fig.~\ref{fig:system}-a1, initially. The layout is intuitive, but it is not space-efficient, as the bottom-right corner is always empty. Although our current design may not be the best choice, it sufficiently meets \rcbox{MC1}. The ML experts commented that the design is easy to understand, and they like the way we metaphorize the joining process through two overlapped rectangles, and inherently, encode the cell size into the width of the overlapped/non-overlapped regions.

\subsection{Feature View}
The \textit{Feature View} (Fig.~\ref{fig:system}, right) serves \textit{Step 6} of LFD by presenting the meta-features from both the TP and FP discriminators through an ``Overview+Details'' exploration. It is designed to meet \rcbox{MC2} and \rcbox{MC3}, and the design starts from the traditional summary-plot~\cite{lundberg2018consistent} (Fig.~\ref{fig:shap}b), where we found two limitations of the plot on its \textit{visualization accuracy} (Sec.~\ref{sec:accuracy}) and \textit{feature importance order} (Sec.~\ref{sec:featureorder}).

\subsubsection{Visualization Accuracy}
\label{sec:accuracy}
\begin{figure}[tbh]
 \centering 
 \includegraphics[width=0.92\columnwidth]{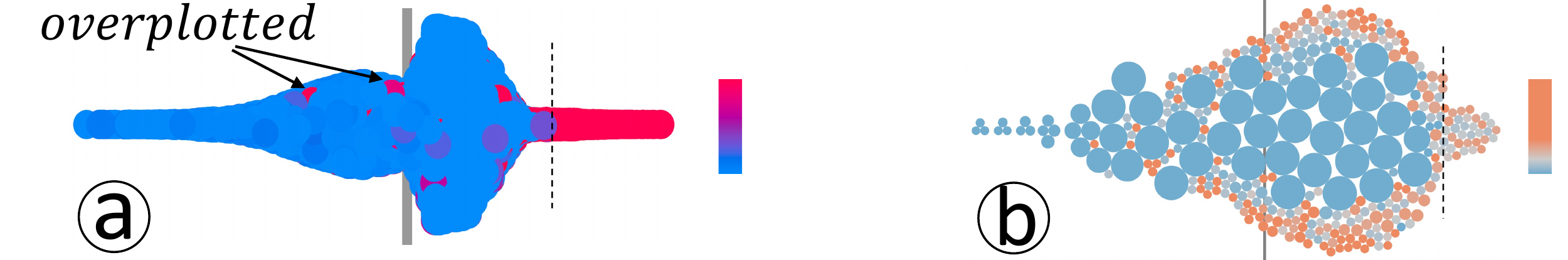}
  \vspace{-0.12in}
 \caption{(a) The overlap of points in a summary-plot may cause misleading visualizations. (b) The visualization on right~\cite{wang2020investigating} avoids overplotting. }
 \label{fig:misleading}
\end{figure}

The accuracy of the summary-plot could be undermined if the number of instances is large. As explained in Fig.~\ref{fig:shap}, each instance is represented as a point. Within the limited space, numerous points will be overplotted and stacked vertically, which may result in misleading visualizations. Fig.~\ref{fig:misleading}a shows a real example (generated using the Python package \texttt{shap}). At first glance, it seems large feature values (the red points on the right of the dashed line) always contribute positively to the prediction. However, there are also red points being plotted \textit{under the blue ones} on the left but are less visible. 
\setlength{\belowcaptionskip}{-8pt}
\setlength{\belowcaptionskip}{0pt}

\begin{figure}[tb]
 \centering 
 \includegraphics[width=0.96\columnwidth]{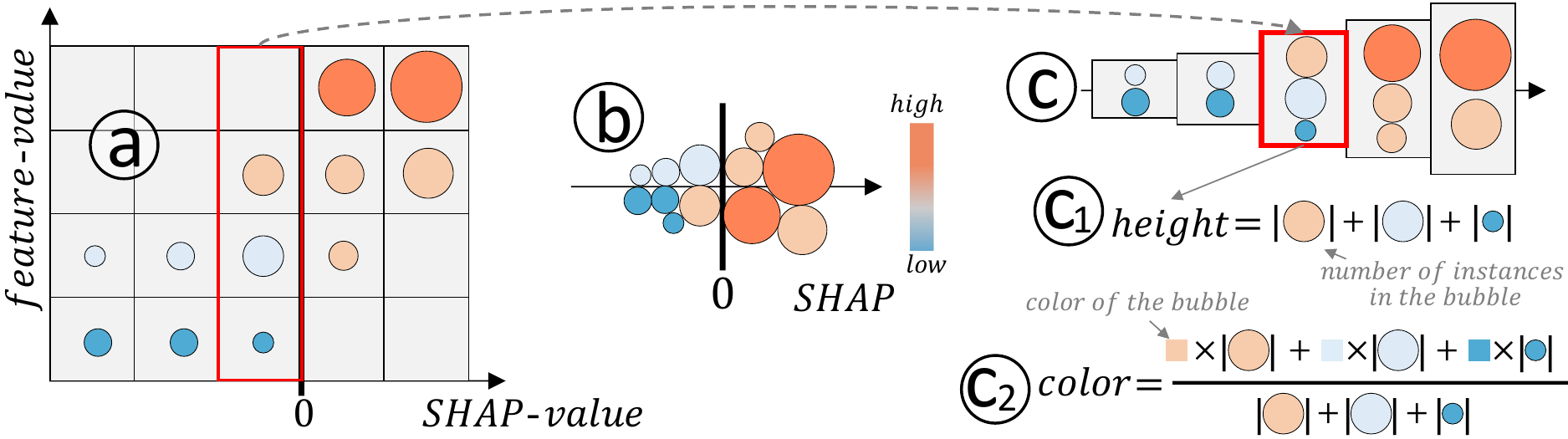}
  \vspace{-0.1in}
 \caption{Visualizing the 2D histogram between feature- and SHAP-values (a) through a group of packed bubbles~\cite{wang2020investigating} (b) or an area-plot (c).}
 \label{fig:bubble}
\end{figure}

This overplotting issue has been discovered by Wang \etal~\cite{wang2020investigating}, and they fixed it by visualizing data distributions as bubbles, rather than individual instances as points. As shown in Fig.~\ref{fig:bubble}a, Wang \etal~\cite{wang2020investigating} construct a 2D histogram of feature- and SHAP-values (i.e., based on the two matrices in Fig.~\ref{fig:shap}a). Non-empty cells of the histogram are represented with bubbles whose size denotes the number of instances. These bubbles are then packed along the $x$-axis (without overlap) based on their SHAP values, using circle packing or force-directed layouts~\cite{zhao2014fluxflow,wang2006visualization,wang2020investigating} (Fig.~\ref{fig:bubble}b). In detail, the circle packing algorithm sequentially places bubbles in tangent to each other while striving to maintain their $x$-position~\cite{zhao2014fluxflow}. Similarly, the force-directed algorithm resolves the overlap among bubbles by iteratively adjusting the bubbles' position, while also trying to retain their $x$-position. The final visualization (Fig.~\ref{fig:misleading}b) resolves the overplotting issue (e.g., many large value instances on the left are visible now). Note that, the number of bubbles is bounded by the product of the number of feature-bins and SHAP-bins in the 2D histogram (Fig.~\ref{fig:bubble}a). Therefore, one can control the number of bubbles by adjusting the number of bins.

The new design, however, cannot accurately reflect the data distribution for two reasons. First, it cannot guarantee to position a bubble at its exact $x$-position, as circles may get shifted to be packed tightly. Second, the size mapping between the number of instances and the bubble size can also cause issues. For example, to make sure bubbles are not too big or too small, size clipping is often applied. If the biggest bubble represents $100$ instances, bubbles with $1000$ instances will have the same size and the accumulated area of bubbles cannot accurately reflect the data distribution.
To fix this, we draw the data distribution as a set of horizontally stacked rectangles (Fig.~\ref{fig:bubble}c), whose height accurately reflects the number of instances in the corresponding SHAP bin (Fig.~\ref{fig:bubble}-c1). The color of a rectangle is blended through a weighted-sum of all bubbles in the SHAP bin (Fig.~\ref{fig:bubble}-c2). Increasing the granularity of the SHAP bins, we get the final area-plot visualization for one feature, e.g., Fig.~\ref{fig:system}-b2.

To guarantee interpretation accuracy (\rfbox{FA1}) and enable interactivity, we employ both designs in Fig.~\ref{fig:bubble}b,~\ref{fig:bubble}c through an ``Overview+Details'' design. All meta-features from the TP and FP discriminators are presented as two columns of area-plots for an overview, where the same features are connected across columns (Fig.~\ref{fig:system}-b5, \rcbox{MC3}). When clicking a meta-feature, its details will be shown using the bubble-plot (Fig.~\ref{fig:system}-b3). We also improve the color legend as an interactive ``transfer function''. Users can add/drag/delete the control-points on the legend to change the color mapping. Brushing the legend will select bubbles with feature values in the brushed range (an example is shown in Fig.~\ref{fig:vda}c). 

\setlength{\belowcaptionskip}{-12pt}
\setlength{\belowcaptionskip}{0pt}

\subsubsection{Feature Importance Order}
\label{sec:featureorder}
In real-world applications, it is very common to work with hundreds of features. Effectively identifying the important ones becomes critical (\rfbox{FA2}). The summary-plot prioritizes features by $mean(|SHAP|)$~\cite{lundberg2018consistent}, which is an effective metric in reflecting the \textit{\textbf{Magnitude}} of features' SHAP values, i.e.,
	\begin{equation}\scriptsize
	Magnitude {=} \frac{1}{n}\sum_{i=1}^{n} |SHAP_i|,\ \ n\ is\ the\ number\ of\ instances.
	\end{equation}

\begin{figure}[tb]
 \centering 
 \includegraphics[width=0.75\columnwidth]{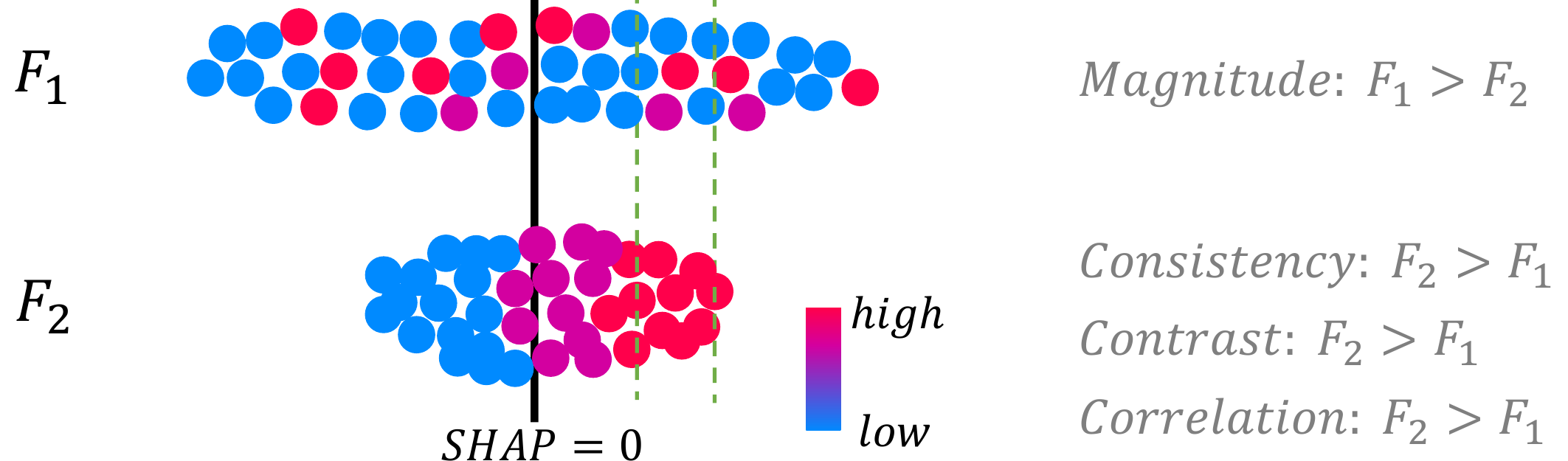}
  \vspace{-0.12in}
 \caption{$F_1$ is more important than $F_2$ as its magnitude is larger. However, $F_2$ is more consistent and more contrast compared to $F_1$.}
 \label{fig:order}
\end{figure}

\begin{figure}[b]
 \centering 
 \includegraphics[width=\columnwidth]{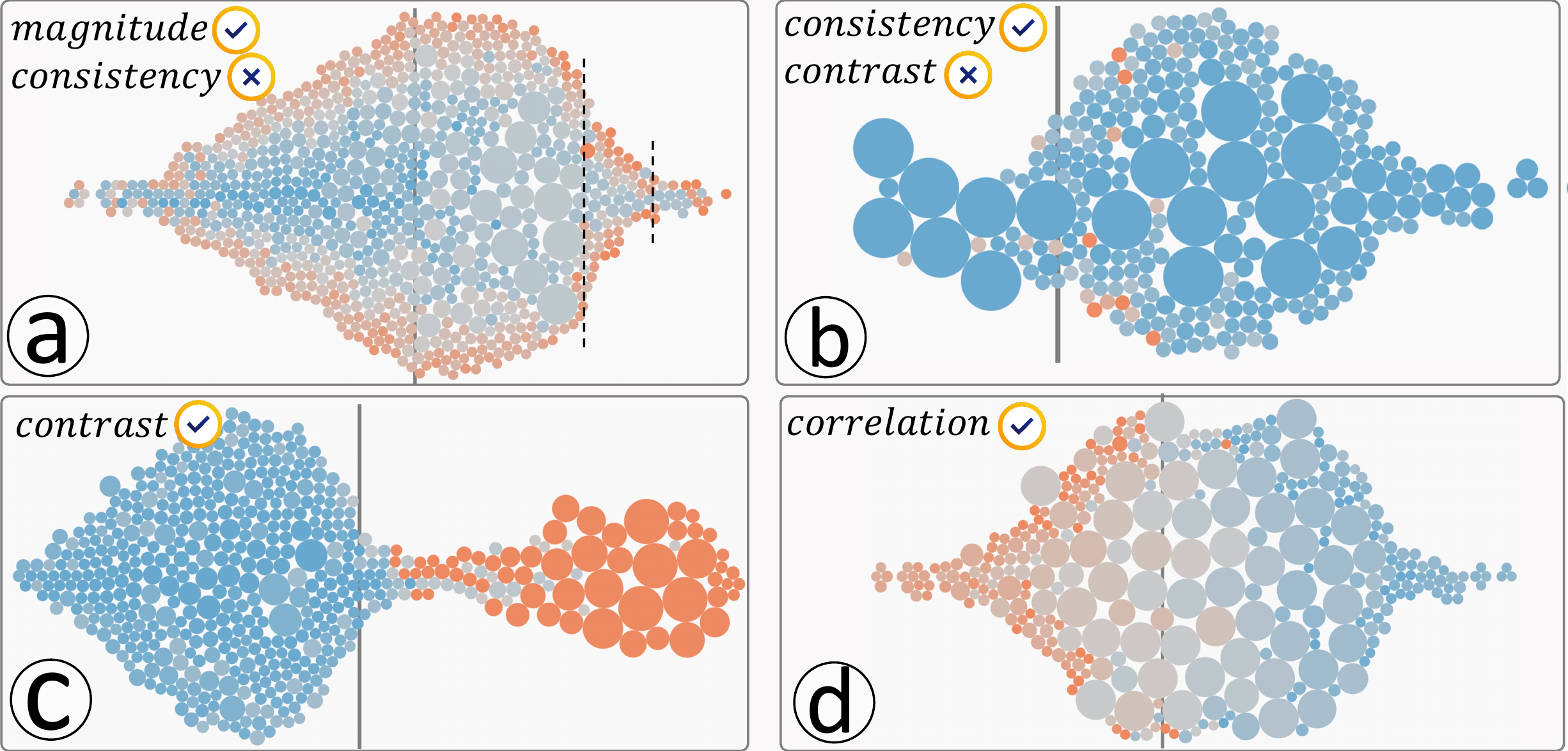}
  \vspace{-0.25in}
 \caption{Important features identified by different metrics: (a) \textit{Magnitude}; (b) \textit{Consistency}; (c) \textit{Contrast}; and (d) \textit{Correlation}.}
 \label{fig:sort}
\end{figure}

However, we found this metric often fails to bring the most important features up, as an example shown in Fig.~\ref{fig:order}. Feature $F_1$ has a larger absolute magnitude than $F_2$, as the points are distributed more widely in the horizontal direction. However, the contribution of $F_1$ is less consistent than $F_2$. For example, within the two dashed lines, both small and large $F_1$ values (blue and red points) could have positive contributions ($SHAP{>}0$) to the final prediction. In contrast, only large $F_2$ values contribute positively. Fig.~\ref{fig:sort}a shows a real large-magnitude feature. Although the feature's contribution magnitude is large, its contribution is not consistent, as the blue and orange bubbles are mixed within any horizontal range. Apparently, measuring features' importance by their magnitude only is not sufficient.

To accommodate the above issue, we propose the \textit{\textbf{Consistency}} metric. It is computed by (1) calculating the entropy of the feature-values in each SHAP bin (a column of cells in Fig.~\ref{fig:bubble}a), (2) summing up the entropy from all SHAP bins, using the number of instances in each bin as the weight, (3) taking the inverse of the sum value. Mathematically, 
	\begin{equation}\scriptsize
	Consistency {=}\Big(\frac{1}{n} \sum_{i=1}^{m} \pazocal{H}\big(F(x), \forall x{\in}bin_i\big){\times}|bin_i|\Big)^{-1},\ n{=}\sum_{i=1}^{m} |bin_i|,
	\end{equation}
\noindent
where $F(x)$ retrieves the feature-value of data instance $x$, $|bin_i|$ denotes the size of a bin, $\pazocal{H}()$ computes the entropy, and $m$ is the number of SHAP bins. 
Fig.~\ref{fig:sort}b presents a real feature with high consistency, reflected by the homogeneous color of bubbles within different SHAP bins (horizontal ranges). However, this feature is not very useful. Because small values (in blue) are largely distributed on both sides of the $SHAP{=}0$ line, and could contribute both positively and negatively. Consequently, solely relying on the consistency metric cannot effectively identify important features either.

\setlength{\belowcaptionskip}{-8pt}
\setlength{\belowcaptionskip}{0pt}

\setlength{\belowcaptionskip}{-12pt}
\setlength{\belowcaptionskip}{0pt}

To capture features with clear contribution differences, we propose the \textit{\textbf{Contrast}} metric, which is computed by the Jensen-Shannon divergence (JSD~\cite{lin1991divergence}) between two normalized distributions formed by the feature-values with positive and non-positive SHAP values, i.e.,
\begin{equation}\scriptsize
\label{eq:contrast}
Contrast {=} JSD\Big(\pazocal{D}\big(F(x) | SHAP_x{<=}0\big)\ {||}\ \pazocal{D}\big(F(x) | SHAP_x{>}0\big)\Big)
\end{equation}
\noindent
where $\pazocal{D}()$ denotes the operation of forming a normalized distribution from a set of values.
Fig.~\ref{fig:sort}c demonstrates a typical contrast feature. This feature is a very good one as it has very clear feature contributions. Large feature values (orange bubbles) always contribute positively and small values (blue bubbles) always contribute negatively.

Lastly, we also compute the absolute Pearson \textit{\textbf{Correlation}} between the SHAP-values and feature-values. This metric further enhances the contrast metric by revealing if the feature-values are linearly correlated with their contributions or not. Fig~\ref{fig:sort}d shows a feature with a large correlation. Smaller feature-values (in darker blue) contribute more positively to the prediction and the contribution is roughly monotonic (i.e., from left to right, the color changes from dark orange, light orange, light blue, to dark blue).

We claim that features' importance should be evaluated from multiple perspectives, and thus, integrate the four metrics with a weighted-sum to generate the fifth metric, i.e., \textit{\textbf{Overall}} (Eq.~\ref{eq:overall}). Note that, the weights here are derived based on preliminary studies with the Avazu dataset~\cite{openctr}. However, they could be different for different datasets.
\begin{equation}\scriptsize
\label{eq:overall}
Overall {=} 2{\times}Magnitude {+} Consistency{+}2{\times}Contrast{+}Correlation
\end{equation}

The TP and FP discriminators have the same set of meta-features, and they are presented as two columns, which can be easily ordered by the five metrics (Fig.~\ref{fig:system}-b5, \rfbox{FA2}). The same features across columns are linked by a curve for tracking (\rcbox{MC3}). The visualization in Fig.~\ref{fig:system}-b6 demonstrates the ranks of the selected meta-feature in all five metrics.

\section{Case Studies}
\label{sec:casestudy}
This section demonstrates cases of using LFD in two real-world applications, (1) merchant category verifications from the payment industry; (2) CTR predictions for advertising. The case studies were conducted with the eight ML experts introduced in Sec.~\ref{sec:requirement}. We used the ``\textit{guided exploration + think-aloud discussions}'' as the protocol to conduct the studies in three steps, (1) explaining the design goals of LFD and our visualizations; (2) guiding the experts to explore the cases, our system, and explain the corresponding findings; (3) open-ended interview to collect their feedback (Sec.~\ref{sec:feedback}).

\subsection{Models for Merchant Category Verification}
\label{sec:vda}

In the financial services industry, each registered merchant has a category reflecting its service type, e.g., \textit{restaurants}, \textit{pharmacies}, and \textit{casino gaming}. The category of a merchant may get misreported for various reasons, e.g., high-risk merchants may report a fake category with lower risk to avoid high processing fees~\cite{yeh2020merchant}. Therefore, it is important to verify the category reported by each merchant. The credit card transactions of a merchant, depicting the characteristic of its provided service, are often used to solve this problem.

For simplicity, we compare binary classifiers for \textit{restaurant} verification only in this work (positive: \textit{restaurant}, negative: \textit{non-restaurant}). Four classifiers have been introduced by the ML experts.  Their details are as follows.
\begin{compactitem}
	\item The \textbf{\textit{Tree}} model is an XGBoost, which consumes data in tabular format (rows: merchants, columns: features).
	\item The \textbf{\textit{CNN}} takes sequential features of individual merchants as input (each sequence denotes the values of a merchant's feature across time). It captures temporal behaviors through 1D convolutions in residual blocks.
	\item The \textbf{\textit{RNN}} also captures the merchants' temporal behaviors, but through gated recurrent units (GRUs).
	\item The \textbf{\textit{GNN}} takes both the temporal and affinity information of a merchant into consideration. The temporal part is managed by 1D convolutions, whereas the affinity part is derived from a graph of merchants. Two merchants are connected if at least one card-holder visited both, and the strength of the connection is proportional to the number of shared card-holders. A \textit{GNN} is then built to learn from this weighted graph of merchants.
\end{compactitem}

More details on the models' architectures can be found from~\cite{yeh2020merchant}. However, LFD does not need these details since it is model-agnostic.
From the training data (raw transactions), the four models use different feature extraction methods to derive their respective training features in different formats, e.g., some are in tabular form and some are in sequences. Even we have little knowledge about the features used by different classifiers, we can still compare them with LFD (i.e., feature-agnostic). For the comparisons, we have $\sim$3.8 million merchants and their raw transactions in 2.5 years.

Since the experts build these models, they have sufficient knowledge on them. We compare the models using LFD and verify the derived insights with the experts for sanity checks in Sec.~\ref{sec:check}. 
Sec.~\ref{sec:cnnrnn} presents new insights to deepen the experts' understanding of the models.

\begin{table*}[tbh]\scriptsize
\centering
\caption{Using LFD to compare the \textit{Tree} and \textit{RNN} models (top row), the \textit{RNN} and \textit{GNN} models (bottom row) for sanity checks (see details in Sec.~\ref{sec:check}). The action items at each step start with ``$\triangleright$'', and the output from each step (which serves as the input to the next step) starts with ``$\Rightarrow$''.}
\label{tab:check}
\vspace{-0.12in}
\begin{tabular}{cccccc}
\steparrow{clrstep1}{1.25}{0.32}{Step 1}& \steparrow{clrstep2}{2.46}{0.32}{Step 2} & 
\steparrow{clrstep3}{2.67}{0.32}{Step 3}& \steparrow{clrstep4}{3.15}{0.32}{Step 4} & 
\steparrow{clrstep5}{2.95}{0.32}{Step 5}& \steparrow{clrstep6}{1.42}{0.32}{Step 6} \\ \hline
\multicolumn{1}{|l|}{\multirow{3}{*}{\begin{tabular}[c]{@{}l@{}}$\triangleright$ \textit{A}: \textit{Tree}\\ $\triangleright$ \textit{B}: \textit{RNN}\\ $\Rightarrow$scores\end{tabular}}} &
  \multicolumn{1}{l|}{\multirow{6}{*}{\begin{tabular}[c]{@{}l@{}}$\triangleright$ sort merchants by their\\ scores from \textit{A} and \textit{B} \\ $\triangleright$ filter out highly scored \\ ones with a threshold\\ {$\Rightarrow$}two sets of merchants,\\ i.e., {$\text{\textit{A}}^\text{+}$} and {$\text{\textit{B}}^\text{+}$}\end{tabular}}} &
  \multicolumn{1}{l|}{\multirow{6}{*}{\begin{tabular}[c]{@{}l@{}}$\triangleright$ join the two sets of \\ merchants into \\ three cells of the \\ disagreement matrix \\ {$\Rightarrow$}three sets of merchants, \\ i.e., {$\text{\textit{A}}^\text{+}\text{\textit{B}}^\text{+}$, $\text{\textit{A}}^\text{+}\text{\textit{B}}^\text{-}$, $\text{\textit{A}}^\text{-}\text{\textit{B}}^\text{+}$}\end{tabular}}} &
  \multicolumn{1}{l|}{\multirow{6}{*}{\begin{tabular}[c]{@{}l@{}}$\triangleright$ divide each set of merchants \\ into two subsets based on \\ merchants' true label\\ {$\Rightarrow$}six sets of merchants, i.e.,\\ {$\text{\textit{A}}^\text{+}\text{\textit{B}}^\text{+}\text{(TP)}$}, {$\text{\textit{A}}^{\text{+}}\text{\textit{B}}^{\text{-}}\text{(TP)}$, $\text{\textit{A}}^\text{-}\text{\textit{B}}^\text{+}\text{(TP)}$,}\\ {$\text{\textit{A}}^\text{+}\text{\textit{B}}^\text{+}\text{(FP)}$, $\text{\textit{A}}^\text{+}\text{\textit{B}}^\text{-}\text{(FP)}$, $\text{\textit{A}}^\text{-}\text{\textit{B}}^\text{+}\text{(FP)}$}\end{tabular}}} &
  \multicolumn{1}{l|}{\multirow{6}{*}{\begin{tabular}[c]{@{}l@{}}$\triangleright$ generate meta-features\\ $\triangleright$ use {$\text{\textit{A}}^\text{+}\text{\textit{B}}^\text{-}\text{(TP)}$} and {$\text{\textit{A}}^\text{-}\text{\textit{B}}^\text{+}\text{(TP)}$} \\ to train the TP discriminator\\ $\triangleright$ use {$\text{\textit{A}}^\text{+}\text{\textit{B}}^\text{-}\text{(FP)}$} and {$\text{\textit{A}}^\text{-}\text{\textit{B}}^\text{+}\text{(FP)}$}\\ to train the FP discriminator\\ {$\Rightarrow$}TP and FP discriminators\end{tabular}}} &
  \multicolumn{1}{l|}{\multirow{3}{*}{\begin{tabular}[c]{@{}l@{}} $\triangleright$ visualizations \\ and insights\\$\Rightarrow$Fig.~\ref{fig:rnn}\end{tabular}}} \\
\multicolumn{1}{|l|}{} &
  \multicolumn{1}{l|}{} &
  \multicolumn{1}{l|}{} &
  \multicolumn{1}{l|}{} &
  \multicolumn{1}{l|}{} &
  \multicolumn{1}{l|}{} \\
\multicolumn{1}{|l|}{} &
  \multicolumn{1}{l|}{} &
  \multicolumn{1}{l|}{} &
  \multicolumn{1}{l|}{} &
  \multicolumn{1}{l|}{} &
  \multicolumn{1}{l|}{} \\ \cline{1-1} \cline{6-6} 
\multicolumn{1}{|l|}{\multirow{3}{*}{\begin{tabular}[c]{@{}l@{}}$\triangleright$ \textit{A}: \textit{RNN}\\ $\triangleright$ \textit{B}: \textit{GNN}\\ $\Rightarrow$scores\end{tabular}}} &
  \multicolumn{1}{l|}{} &
  \multicolumn{1}{l|}{} &
  \multicolumn{1}{l|}{} &
  \multicolumn{1}{l|}{} &
  \multicolumn{1}{l|}{\multirow{3}{*}{\begin{tabular}[c]{@{}l@{}}$\triangleright$ visualizations\\ and insights \\$\Rightarrow$Fig.~\ref{fig:gnn}\end{tabular}}} \\
\multicolumn{1}{|l|}{} &
  \multicolumn{1}{l|}{} &
  \multicolumn{1}{l|}{} &
  \multicolumn{1}{l|}{} &
  \multicolumn{1}{l|}{} &
  \multicolumn{1}{l|}{} \\
\multicolumn{1}{|l|}{} &
  \multicolumn{1}{l|}{} &
  \multicolumn{1}{l|}{} &
  \multicolumn{1}{l|}{} &
  \multicolumn{1}{l|}{} &
  \multicolumn{1}{l|}{} \\ \hline
\end{tabular}
\end{table*}

\subsubsection{Sanity Checks: [Tree v.s. RNN] and [RNN v.s. GNN]}
\label{sec:check}
We first compare \textbf{the \textit{Tree} (\textit{A}) and \textit{RNN} (\textit{B})}. The \textit{Tree} has a higher AUC (it outperforms the \textit{RNN}), but the \textit{RNN} has a lower LogLoss (the \textit{RNN} outperforms the \textit{Tree}). The performance reflected by the two metrics conflicts with each other, and it is hard to choose the models based on them. Also, the metrics reveal nothing about the models' behavior discrepancy on different features and feature conditions.

Tab.~\ref{tab:check} (top row) shows the LFD comparison process for the two models. \textit{Step 1} feeds the $\sim$3.8 million merchants to the two models and generates two scores for each merchant. \textit{Step 2} uses these scores to sort the merchants decreasingly. Based on a given threshold, we filter out the merchants being predicted as \textit{restaurant}s (i.e., captured) by model \textit{A} and \textit{B} respectively, i.e., $\text{\textit{A}}^\text{+}$ and $\text{\textit{B}}^\text{+}$. \textit{Step 3} joins these two sets and separates the merchants into $\text{\textit{A}}^\text{+}\text{\textit{B}}^\text{+}$, $\text{\textit{A}}^\text{+}\text{\textit{B}}^\text{-}$, and $\text{\textit{A}}^\text{-}\text{\textit{B}}^\text{+}$. Based on the merchants' true label, each cell can further be divided into two smaller cells at \textit{Step 4} (i.e., six cells in total).

\textit{Step 5} generates 70 meta-features for each merchant from its raw transactions over the past 2.5 years. It is not necessary to explain all of them, but the ones used in this case and mentioned later in Sec.~\ref{sec:cnnrnn} include:
\begin{compactitem}
  \item \textit{nonzero\_numApprTrans}: the number of days that a merchant has at least one approved transaction in the 2.5 years. For time-series data, this meta-feature reflects the number of meaningful points in a sequence. It is derived based on the prior knowledge that \textit{RNN}s often behave better on instances with richer sequential information. Its value range is [0, 912] (2.5 years=912 days), reflecting the active level of a merchant.
  \item \textit{nonzero\_amtDeclTrans}: the number of days that a merchant has at least one declined transaction (with a nonzero dollar amount) in the 2.5 years.
  \item \textit{mean\_avgAmtAppr}: the mean of the average approved amount per day, over the 2.5 years, for each merchant.
  \item \textit{mean\_rateTxnAppr}: the mean of the daily transaction approval rate, over the 2.5 years, for each merchant.
  \item \textit{mean\_numApprTrans}: the mean of the number of daily approved transactions, over the 2.5 years, per merchant.
\end{compactitem}
Using all 70 meta-features of the merchants in the $\text{\textit{A}}^\text{+}\text{\textit{B}}^\text{-}$ and $\text{\textit{A}}^\text{-}\text{\textit{B}}^\text{+}$ cells from the TP side, we train the TP discriminator. The FP discriminator is trained using the same meta-features, but the merchants from the corresponding FP cells.

\begin{figure}[tb]
 \centering 
 \vspace{-0.1in}
 \includegraphics[width=\columnwidth]{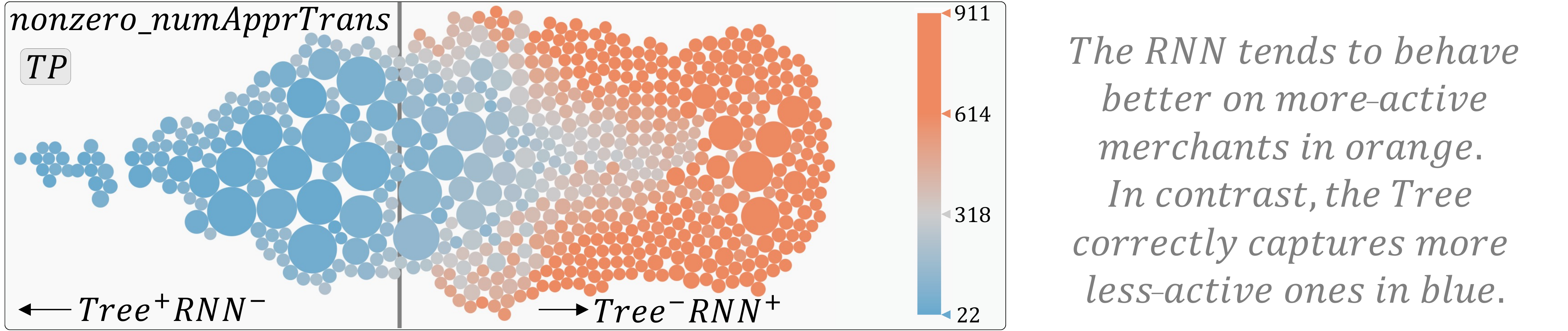}
  \vspace{-0.22in}
 \caption{Compare \textit{Tree} v.s. \textit{RNN}, meta-feature: \textit{nonzero\_numApprTrans}.}
 \label{fig:rnn}
\end{figure}

\textit{Step 6} interprets the discriminators to derive insights. For example, 
from the visualization of \textit{nonzero\_numApprTrans} in the TP side (Fig.~\ref{fig:rnn}), we notice that active merchants in orange bubbles are more likely to be from the $\text{\textit{Tree}}^\text{-}\text{\textit{RNN}}^\text{+}$ (i.e., $\text{\textit{A}}^\text{-}\text{\textit{B}}^\text{+}(\text{TP})$) cell, indicating they will be correctly recognized (captured) by the \textit{RNN}, but missed by the \textit{Tree} (i.e., the \textit{RNN} outperforms the \textit{Tree} on active merchants). 
In contrast, the less active merchants in blue bubbles are more likely to be from the $\text{\textit{Tree}}^\text{+}\text{\textit{RNN}}^\text{-}$ (i.e., $\text{\textit{A}}^\text{+}\text{\textit{B}}^\text{-} (\text{TP})$) cell, indicating the \textit{Tree} outperforms the \textit{RNN} in correctly recognizing less active merchants. The insights here directly guide model selections based on the merchants' service frequency, i.e., active or not.

\setlength{\belowcaptionskip}{0pt}
\setlength{\belowcaptionskip}{0pt}

Next, we compare \textbf{the \textit{RNN} (\textit{A}) and \textit{GNN} (\textit{B})} (Tab.~\ref{tab:check}, bottom row). \textit{Step 1$\sim$4} of the comparison are conducted very similarly to the previous case. The only difference is that we sampled around $\sim$70K merchants from the $\sim$3.8M for this comparison to reduce the merchant graph construction cost.

\begin{figure}[tb]
 \centering 
 \vspace{-0.1in}
 \includegraphics[width=\columnwidth]{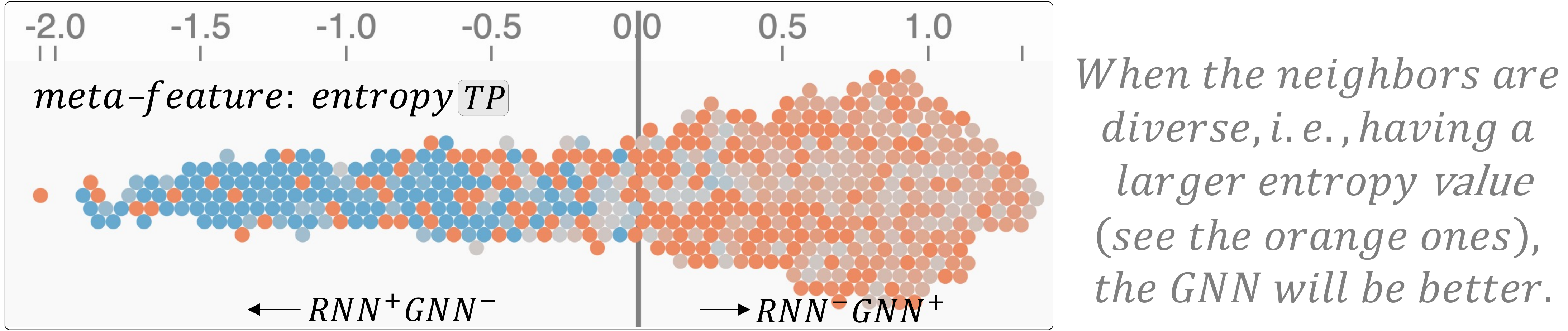}
  \vspace{-0.22in}
 \caption{Compare the \textit{RNN} with \textit{GNN}, meta-feature: \textit{entropy}.}
 \label{fig:gnn}
\end{figure}

At \textit{Step 5}, we generate several affinity-related meta-features to probe the two models, because the experts expect the \textit{GNN} to leverage the information from a merchant's neighbors more than the \textit{RNN}. For example, the \textit{entropy} of a merchant reflects how diverse its neighbors' category is. The \textit{n\_connection} denotes the degree of each merchant in the merchant graph. Using these new meta-features, we train the two discriminators of LFD. From the visualization at \textit{Step 6}, we found that \textit{entropy} is a very differentiable meta-feature (Fig.~\ref{fig:gnn}). When a merchant has more diverse neighbors (i.e., from the orange bubbles with larger \textit{entropy}), the \textit{GNN} tends to correctly capture it whereas the \textit{RNN} is more likely to miss it (i.e., the orange bubbles mostly fall into the $\text{\textit{RNN}}^\text{-}\text{\textit{GNN}}^\text{+}\text{(TP)}$ cell). This indicates the \textit{GNN} outperforms \textit{RNN} on these merchants and the neighbors' information indeed contributes to the predictions. The observation here clearly reveals the value of affinity information between merchants, verifying the improvement from the \textit{GNN}~\cite{yeh2020merchant}.

These two brief cases use prior knowledge on the compared models to derive meta-features. The models' behaviors on the meta-features, in turn, verify the experts' expectations, i.e., our sanity checks succeeded.

\subsubsection{Deeper Insights: [CNN v.s. RNN]}
\label{sec:cnnrnn}
This section presents deeper insights when comparing \textbf{the \textit{CNN} (\textit{A}) and \textit{RNN} (\textit{B})} using LFD. The insights go beyond what was known by the experts and deepen their understandings of the models. The \textit{CNN} and \textit{RNN} have similar performance and both capture the temporal behaviors of the merchants, but in different ways. The \textit{CNN} uses 1D convolutions, whereas the \textit{RNN} employs the GRU structure.

\steparrowfour{clrstep1}{1.2}{0.38}{Step 1}{clrstep2}{2}{clrstep3}{clrstep4}Feeding the $\sim$3.8M merchants to the two models (i.e., two black-boxes), we get two sets of scores, which are used to sort the merchants and identify the two sets of merchants captured by individual models (i.e., $\text{\textit{A}}^\text{+}$ and $\text{\textit{B}}^\text{+}$). Joining the two sets, we get merchants in the three cells of the disagreement matrix (i.e., $\text{\textit{A}}^\text{+}\text{\textit{B}}^\text{+}$, $\text{\textit{A}}^{\text{+}}\text{\textit{B}}^{\text{-}}$, and $\text{\textit{A}}^\text{-}\text{\textit{B}}^\text{+}$). This matrix is then divided into two matrices of six cells based on the merchants' true category label, i.e., $\text{\textit{A}}^\text{+}\text{\textit{B}}^\text{+}\text{(TP)}$, $\text{\textit{A}}^{\text{+}}\text{\textit{B}}^{\text{-}}\text{(TP)}$, $\text{\textit{A}}^\text{-}\text{\textit{B}}^\text{+}\text{(TP)}$, $\text{\textit{A}}^\text{+}\text{\textit{B}}^\text{+}\text{(FP)}$, $\text{\textit{A}}^\text{+}\text{\textit{B}}^\text{-}\text{(FP)}$, and $\text{\textit{A}}^\text{-}\text{\textit{B}}^\text{+}\text{(FP)}$.

\steparrow{clrstep5}{1.2}{0.38}{Step 5}For the ``learning'' part of LFD, we use the merchants from the $\text{\textit{A}}^{\text{+}}\text{\textit{B}}^{\text{-}}\text{(TP)}$ and $\text{\textit{A}}^\text{-}\text{\textit{B}}^\text{+}\text{(TP)}$ cells to train the TP discriminator; the merchants from $\text{\textit{A}}^{\text{+}}\text{\textit{B}}^{\text{-}}\text{(FP)}$ and $\text{\textit{A}}^\text{-}\text{\textit{B}}^\text{+}\text{(FP)}$ to train the FP discriminator. For meta-features, we continue to use the 70 meta-features derived when comparing the \textit{Tree} and \textit{RNN} models in Sec.~\ref{sec:check}.

\setlength{\belowcaptionskip}{0pt}
\setlength{\belowcaptionskip}{0pt}

\setlength{\belowcaptionskip}{-6pt}
\begin{figure*}[tbh]
 \centering 
 \includegraphics[width=\textwidth]{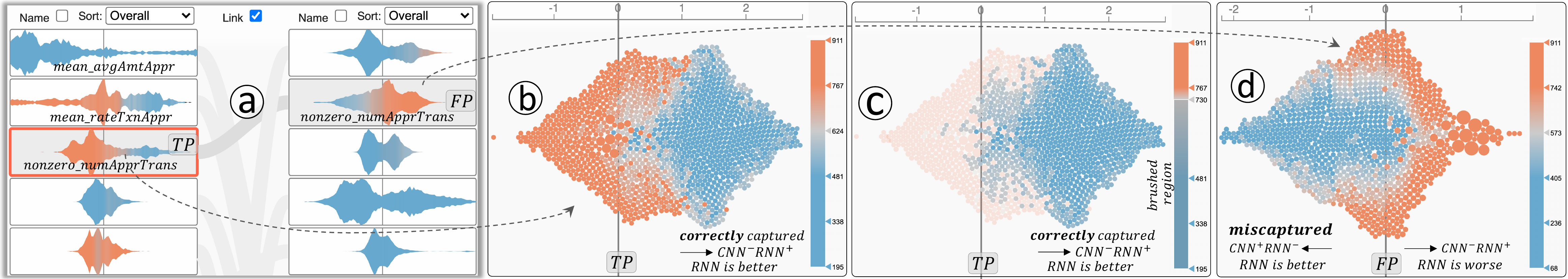}
  \vspace{-0.27in}
 \caption{Compare the \textit{CNN} and \textit{RNN}. (a) TP (left) and FP (right) meta-feature lists. (b) \textit{nonzero\_numApprTrans} ranks the third in the TP list. The \textit{RNN} performs better when this meta-feature has a smaller value. (c) is the same with (b) but highlights merchants that are less active than 2 years (730 days). (d) \textit{nonzero\_numApprTrans} on the FP side shows a reversed pattern with (b). All instances here are \textit{non-restaurant}s. The \textit{RNN} still behaves better than \textit{CNN} on less-active merchants, i.e., the \textit{RNN} will NOT mis-capture less-active merchants (the blue bubbles on the left).}
 \label{fig:vda}
\end{figure*}
\setlength{\belowcaptionskip}{0pt}

\setlength{\belowcaptionskip}{0pt}\setlength{\belowcaptionskip}{-6pt}
\begin{figure*}[tbh]
 \centering 
 \includegraphics[width=\textwidth]{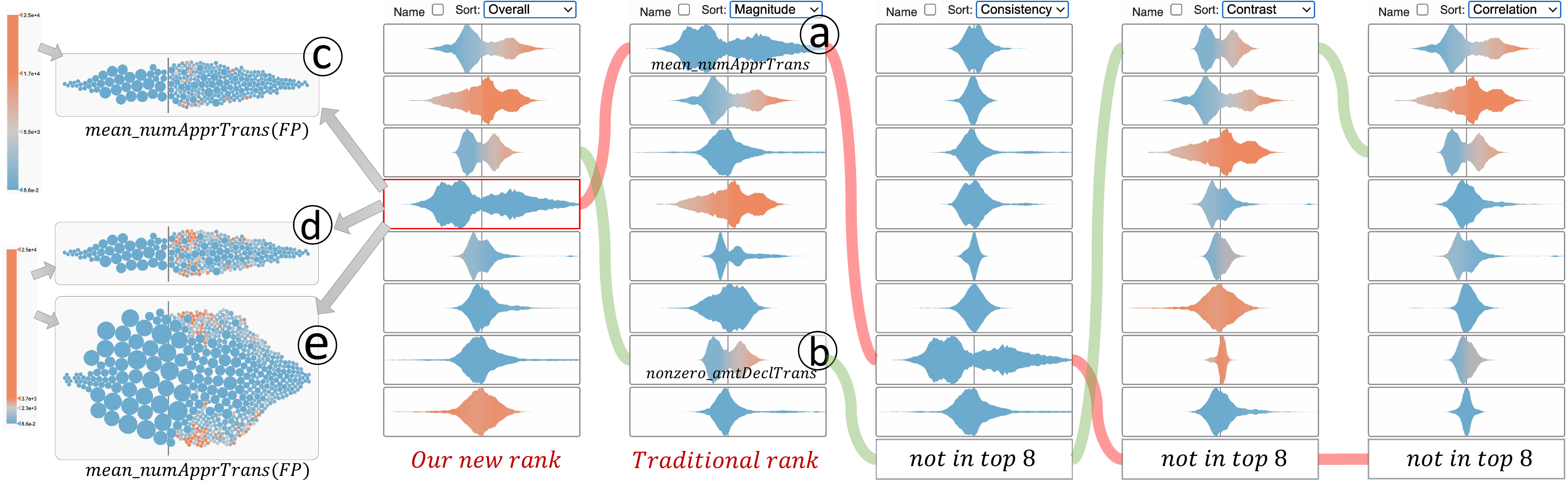}
  \vspace{-0.3in}
 \caption{The ranks of meta-features on the FP side (top-8 only). (a) \textit{mean\_numApprTrans} shows large \textit{Magnitude} but very small \textit{Contrast} and \textit{Correlation} (not in the top-8), thus its \textit{Overall} ranking becomes lower. (b) \textit{nonzero\_amtDeclTrans} ranks the 7th in \textit{Magnitude}. The good \textit{Contrast} and \textit{Correlation} bring it up in the  \textit{Overall} ranking. (c, d) The same summary-plot with different color mappings. (d, e) The same summary-plot with different granularities, i.e., different numbers of bins in the 2D histogram of feature- and SHAP-values (details explained in Fig.~\ref{fig:bubble}a).}
 \label{fig:rank}
\end{figure*}
\setlength{\belowcaptionskip}{0pt}

\steparrow{clrstep6}{1.2}{0.38}{Step 6}Fig.~\ref{fig:vda}a shows the important meta-features to the TP and FP discriminators, both sides are ranked by the \textit{Overall} metric (Eq.~\ref{eq:overall}, \rcbox{MC3}). On the TP side (left), \textit{nonzero\_numApprTrans} ranks the third and its detail is shown in Fig.~\ref{fig:vda}b. The \textit{RNN} correctly captures more when the merchants are relatively less-active (i.e., the blue bubbles on right are more likely to be $\text{\textit{CNN}}^\text{-}\text{\textit{RNN}}^\text{+}$), whereas the \textit{CNN} correctly captures more when the merchants are very active in the past 2.5 years (the orange bubbles). With some analysis, the models' behavior matched the experts' expectations and they commented that the \textit{CNN} has a limited receptive field (limited by the number of convolutional layers), and focuses more on local patterns, whereas the \textit{RNN} can memorize longer history through its internal hidden states. When a merchant has a long and active history, there are rich local patterns as well, and the \textit{CNN} outperforms the \textit{RNN}. However, when the merchant is less-active, the \textit{RNN} performs better in capturing the \textit{sparser} temporal behaviors.
Fig.~\ref{fig:vda}c shows the same result with user-interactions, i.e., by dragging the color control-points and brushing the legend, we can easily identify merchants that the \textit{RNN} outperforms the \textit{CNN} (i.e., the ones with approved transactions less than 730 days). Using this meta-feature, LFD successfully reveals the subtle difference between the two models (\rcbox{MC2}, \rfbox{FA1}).

Another meta-feature, i.e., \textit{mean\_rateTxnAppr} (ranked the second on the TP side), demonstrates the same trend, indicating the \textit{RNN} outperforms \textit{CNN} in correctly identifying \textit{restaurant}s with lower approval rates as well. In contrast, the \textit{mean\_avgAmtAppr} (ranked the first) shows unnoticeable difference for the merchants on the two sides of the $SHAP{=}0$ line (the areas on both sides are in blue), implying this meta-feature cannot differentiate the two models. However, the large magnitude of this feature still makes it rank the first.

On the FP side (Fig.~\ref{fig:vda}a, right), \textit{nonzero\_numApprTrans} ranks the second, and its detail is shown in Fig.~\ref{fig:vda}d. All merchants here are \textit{non-restaurant}s but mis-captured as \textit{restaurant}s (i.e., FP), and the bubbles' pattern is reversed comparing to Fig.~\ref{fig:vda}b (i.e., the blue bubbles come to the left in Fig.~\ref{fig:vda}d). 
The less-active merchants in blue bubbles are less likely to be mis-captured by the \textit{RNN}, but more likely to be mis-captured by the \textit{CNN} (i.e., from the $\text{\textit{CNN}}^\text{+}\text{\textit{RNN}}^\text{-}$ cell of the FP side), indicating the \textit{RNN} still outperforms the \textit{CNN} on merchants with \textit{sparser} temporal behaviors.

We have also explored different metrics in ranking the meta-features. Fig.~\ref{fig:rank} shows the top-8 features ranked on the FP side. \textit{mean\_numApprTrans} (Fig.~\ref{fig:rank}a) has the largest \textit{Magnitude} and it is identified as the most important meta-feature using the traditional order (i.e., $mean(|SHAP|)$). However, it ranks the 7th in the \textit{Consistency} order and is not in the top-8 in the \textit{Contrast} and \textit{Correlation} orders (tracking the red curve). 
In contrast, \textit{nonzero\_amtDeclTrans} (Fig.~\ref{fig:rank}b, tracking the green curve) is the 7th important in the \textit{Magnitude} list but has very large \textit{Contrast} and \textit{Correlation} values. Compared to the traditional \textit{Magnitude} order, our \textit{Overall} metric improves the rank of \textit{nonzero\_amtDeclTrans} (ranked the 3rd) and decreases the rank of \textit{mean\_numApprTrans} (ranked the 4th), by considering multiple aspects of the meta-features (\rfbox{FA2}).

Note that, the visual appearance of the area-plot and bubble-plot depends on the color mapping, which can be adjusted by the ``transfer function'' widget.  
Fig.~\ref{fig:rank}c shows the detailed view of \textit{mean\_numApprTrans} with the default color legend. It can be seen that there are very few orange bubbles and their contribution to the aggregated color of the area-plot is unnoticeable. We can change the color mapping, as shown by the legend in Fig.~\ref{fig:rank}d, to map a larger range of values to orange. However, most of the bubbles are still in blue, and the meta-feature still has small \textit{Contrast}.

The layout and size of the bubbles can also be flexibly adjusted to reflect different levels of details for a meta-feature. Fig.~\ref{fig:rank}d-e demonstrate this by increasing the number of feature- and SHAP-bins when computing the 2D histogram (explained in Fig.~\ref{fig:bubble}a), e.g., Fig.~\ref{fig:rank}e uses more bins and presents the feature in finer granularity. This also reflects that the accumulated area of the bubble-plot (the \textit{Detail} part) cannot accurately present the data distribution, and verifies the need of the area-plot (the \textit{Overview} part), see Sec.~\ref{sec:accuracy}.

\subsection{Models for CTR Prediction}
\label{sec:ctr}
CTR prediction is to predict if a user will click an advertisement or not, which is a critical problem in the advertising industry. Avazu~\cite{openctr} is a public CTR dataset, containing more than 40 million instances across 10 days. Each instance is an impression (advertisement view) and has 21 anonymized categorical features, e.g., \textit{site\_id}, \textit{device\_id} (see details in~\cite{avazufeat}).

\steparrow{clrstep1}{1.2}{0.38}{Step 1}Using Avazu, we compare a tree-based model ($\text{\textit{A}}$, short for \textbf{\textit{Tree}}) and an \textbf{\textit{RNN}} model ($\text{\textit{B}}$). Both are trained to differentiate \textit{click} from \textit{non-click} instances. The following data partition was used in both models' training:
\begin{compactitem}
	\item Day 1: reserved to generate historical features.
	\item Day 2-8: used for model training
	\item Day 9: used for model testing and comparison
	\item Day 10: held-out for model ensembling experiments
\end{compactitem}
As the models may use some historical features, e.g., the number of active hours in the past day, we reserve Day 1 data for this purpose. Also, we do not touch Day 10 data, and leave it for our later quantitative evaluation experiments. Note that, there are also works that partition Avazu by shuffling the 10 days data~\cite{li2019fi} and dividing them into folds. Our way of partitioning data by time follows our industrial practice and is more realistic, i.e., we should not leak any future data into the training process.

The \textit{Tree} model takes individual data instances (advertisement views) as input, whereas the \textit{RNN} connects instances into viewing sequences and takes them as input. We followed the winning solution of the \textit{Avazu CTR Challenge} to form the sequences~\cite{rnnid}. The details of individual models' training features and architectures are not needed, as LFD is feature-agnostic and model-agnostic. The final AUCs of the \textit{Tree} and \textit{RNN} are 0.7468 and 0.7396 respectively.

\steparrowthree{clrstep2}{1.2}{0.38}{Step 2}{clrstep3}{3}{clrstep4}These steps are similar to what we have described in earlier cases. Fig.~\ref{fig:system} (left) shows the \textit{Disagreement Distribution View} for this case, from which, we know that the size of the disagreed instances from the TP side reaches the peak if the cutoff/threshold is between 15\%$\sim$20\% (Fig~\ref{fig:system}-a3). We use 15\% as the cutoff to maximize the size of training data with the most disagreed predictions (\rcbox{MC1}). One can also choose other cutoffs based on the FP data distributions. However, we care more about TP instances in this case.

\steparrow{clrstep5}{1.2}{0.38}{Step 5}This step generates meta-features from the raw data in two steps. First, the Avazu raw data has 21 categorical features, and we extend them by concatenating the features' categorical values, e.g., \textit{device\_id\_ip} is a feature combined from \textit{device\_id} and \textit{device\_ip}. Based on the experts' knowledge on the data, we extend Avazu to have 42 features (and feature combinations). Second, as CTR is the ratio between the number of clicks (\textit{n\_clicks}) and impressions (\textit{n\_impressions}), i.e., \textit{CTR=n\_clicks/n\_impressions}, we propose meta-features by profiling the frequency of the 42 features from these three dimensions. For example, for the raw feature \textit{device\_id}, we generate meta-feature \textit{\textbf{n\_clicks\_}device\_id}, denoting the number of clicks per \textit{device\_id} value. Also, as the \textit{RNN} is not as good as the \textit{Tree} (see the AUCs at \textit{Step 1}), we want to probe the models' sequence-related behaviors. Thus, we generate meta-features from a fourth dimension to reflect the active level of the features (i.e., \textit{\textbf{n\_active\_hours\_}*}). In summary, generating meta-features from the 42 features in the following four dimensions, we get 168 ($42{\times}4$) meta-features (their details are included in our Appendix),
\begin{compactitem}
	\item \textit{\textbf{n\_impressions\_}*} denotes the number of impressions per value of * (* represents one of the 42 features).

	\item \textit{\textbf{n\_clicks\_}*} indicates the number of clicks per value of *.
	
	\item \textit{\textbf{ctr\_}*} denotes the CTR for each value of *. For example, if * is \textit{device\_id} and the CTR for a certain device is $x$, then the value of \textit{\textbf{ctr\_}device\_id} will be $x$ for all impressions happened on this device.
	
	\item \textit{\textbf{n\_active\_hours\_}*} reflects the number of hours (in the past day) that each value of * appeared in the data.
\end{compactitem}
Using the 168 meta-features, we train the TP and FP discriminators to learn the difference between the \textit{Tree} and \textit{RNN}.

\steparrow{clrstep6}{1.2}{0.38}{Step 6}After training, we use our LFD system to rank the meta-features from the TP and FP sides (Fig.~\ref{fig:system}-b5). Meta-features \textit{\textbf{n\_active\_hours\_}*} never appear in the top differentiable ones, implying the \textit{RNN} may not benefit more from the sequential information (compared to the \textit{Tree}). This insight provides clues to further diagnose the \textit{RNN} (\rcbox{MC2}).

After some analysis, the ML experts tend to believe the relatively worse performance of \textit{RNN} is from the data. \textit{First}, there is no unique sequence id in Avazu, and the solution we adopted to connect instances into sequences (from~\cite{rnnid}) may not be optimal. \textit{Second}, the sequence length is very short and the only 10 days data may not be able to well capture users' temporal behaviors (detailed sequence statistics are in our Appendix). In contrast, in Sec.~\ref{sec:vda}, we used 2.5 years of data and each merchant has a unique sequence id.
The interpretations provide some explanations on why \textit{RNN} is not one of the top solutions for this CTR challenge~\cite{avazurank}.

From the TP side (Fig.~\ref{fig:system}-b5, left), we can see \textit{\textbf{ctr\_}site\_id} is a very differentiable feature. Although the \textit{RNN} has a smaller AUC, it outperforms the \textit{Tree} in capturing \textit{clicks} when this meta-feature is large, i.e., the orange bubbles in Fig.~\ref{fig:system}-b3 are more likely to be from the $\text{\textit{Tree}}^\text{-}\text{\textit{RNN}}^\text{+}$ cell. This makes sense as the \textit{RNN} tends to remember the site visiting history. From the meta-features ordered by the \textit{Overall} metric, we can easily identify other important ones. For example, \textit{\textbf{ctr\_}site\_app\_id} (ranked the second) denotes the CTR of the feature combined from \textit{site\_id} and \textit{app\_id}. It shows the same trend with the meta-feature \textit{\textbf{ctr\_}site\_id}, i.e., the \textit{RNN} behaves better than the \textit{Tree} if an impression is from the site-application pair with a higher click rate. \textit{\textbf{ctr\_}c14} (ranked the third) is another important meta-feature. The \textit{RNN} would be more accurate in capturing \textit{clicks} when this feature has small values (i.e., indicated by the blue area on the right of Fig.~\ref{fig:system}-b7). Although we do not know what \textit{c14} is (an anonymized raw feature of Avazu), we can infer it is an important click-related feature from LFD.

For the FP side (Fig.~\ref{fig:system}-b5, right), \textit{\textbf{ctr\_}site\_id} is also the most important (Fig.~\ref{fig:system}-b4, tracking the curves between the two meta-feature lists, \rcbox{MC3}). Its behavior is consistent to the TP side (Fig.~\ref{fig:system}-b2), i.e., orange regions are still on the right, indicating the \textit{RNN} tends to mis-capture instances if the value of \textit{\textbf{ctr\_}site\_id} is large. In contrast, the good feature we saw from Sec.~\ref{sec:cnnrnn} (i.e., \textit{nonzero\_numApprTrans}) shows reversed patterns from the TP and FP sides (Fig.~\ref{fig:vda}b,~\ref{fig:vda}d). This reveals a potential bias of the model, i.e., the \textit{RNN} here tends to give high scores to instances with large \textit{\textbf{ctr\_}site\_id}, \textbf{no matter} they are real \textit{click} instances or not. After thoroughly examining and comparing the two cases, we found that the observation actually reflects different \textit{signal strengths} of the merchant category verification and CTR problems.

As explained by the ML experts, the \textit{signal strength} of a dataset reflects how distinguishable the positive instances are from the negative ones. For merchant category verification, the signal strength is very strong, i.e., merchants with falsified categories have certain on-purpose behaviors that regular merchants cannot accidentally conduct. However, for CTR, the signal strength is much weaker. Randomness widely exists when users choose to click or not. As a result, two records with similar values across all features may have different \textit{click} labels. Consequently, some instances from the $\text{\textit{A}}^\text{-}\text{\textit{B}}^\text{+} (\text{TP})$ and $\text{\textit{A}}^\text{-}\text{\textit{B}}^\text{+} (\text{FP})$ cells are similar, so as to some instances from the $\text{\textit{A}}^\text{+}\text{\textit{B}}^\text{-} (\text{TP})$ and $\text{\textit{A}}^\text{+}\text{\textit{B}}^\text{-} (\text{FP})$ cells. In turn, the trained TP and FP discriminators behave similarly (due to their similar training data), which explains the similar patterns of \textit{\textbf{ctr\_}site\_id} on the TP and FP sides (Fig.~\ref{fig:system}-b2, b4).

\section{Evaluation and Feedback}
\label{sec:evaluation}

This section quantitatively evaluates the feature-ordering metrics (using FWLS) and presents an ablation study to profile how the number and quality of meta-features affect the training of the discriminators at \textit{Step 5} of LFD. Also, we summarize the feedback and suggestions provided by the ML experts during the open-ended interviews with them.

\subsection{Quantitative Evaluation with Model Ensembling}
\label{sec:ensemble}
As explained in Sec.~\ref{sec:bgensemble}, FWLS ensembles models by considering the models' behaviors in different features. What features should be used in FWLS is a critical problem and the ensembling result can quantitatively reflect the quality of the used features. We use the top-15 meta-features ranked by the five metrics of LFD to generate five FWLS models, and compare their performance to quantify the rankings' quality. Note that, the ensembling experiments here can be fully automatic and do not rely on any visual inputs. However, to be consistent with our earlier writing style and simplify the logic, we still describe them following the six steps of LFD and explain what is needed/expected in individual steps.

\steparrow{clrstep1}{1.2}{0.38}{Step 1}We use two state-of-the-art and publicly available CTR models to conduct this experiment, so that people can reproduce it. One is the logistic regression (\textbf{\textit{LR}}) implemented by ``follow the proximally regularized leader (FPRL-Proximal)''~\cite{mcmahan2013ad}. The other is the ``feature interaction graph neural network'' (Fi-GNN, short for \textbf{\textit{GNN}})~\cite{li2019fi}, currently ranking the first for this challenge~\cite{avazurank}. Both models are trained using the 21 raw features of Avazu (Day 2$\sim$8).

\steparrowthreegap{clrstep2}{1.2}{0.38}{Step 2}{clrstep3}{3}{clrstep5}{5}Using scores of the two models, we sort data instances (i.e., test data from Day 9), identify the ones in the $\text{\textit{LR}}^\text{+}\text{\textit{GNN}}^\text{-}$ and $\text{\textit{LR}}^\text{-}\text{\textit{GNN}}^\text{+}$ cell (\textit{Step 3}), and train a discriminator to differentiate them using the 168 meta-features proposed in Sec.~\ref{sec:ctr} (\textit{Step 5}). These meta-features have no overlap with the training features of the \textit{LR} or \textit{GNN}. Note that, \textit{Step 4} of LFD is not needed here, since we need a single order of meta-features considering both TP and FP instances, i.e., no need to separate the TP and FP instances.

\steparrow{clrstep6}{1.2}{0.38}{Step 6}After getting the discriminator, we use the five metrics from Sec.~\ref{sec:featureorder} to rank the 168 meta-features into five orders, and pick the top-15 from each to conduct a FWLS. As a result, there will be five models ensembled from the original \textit{LR} and \textit{GNN}, each uses a set of 15 meta-features that are considered as the most important by one metric. The performance of the five FWLS models reflects how useful the five sets of meta-features are, which in turn, reflects the quality of the five metrics. All the FWLS models are trained on Day 9 data to learn the values of $w_1^i$ and $w_2^i$ in Eq.~\ref{eq:fwls}, and we test their performance on both Day 9 and Day 10 data.

\setlength{\textfloatsep}{0.12cm}
\begin{table}[b]\scriptsize
\caption{Performance of the \boxsingle{original} \textit{LR}, \textit{GNN}, and the five models \boxensemble{ensembled} from them (using the top-15 meta-features from different metrics).}
\vspace{-0.12in}
\begin{tabular}{|m{.65in}|m{.54in}|m{.48in}|m{.408in}|p{.52in}|}
\hline
\cellcolor{gray!25} \textit{Model}&\cellcolor{gray!25}\textit{AUC (Day 9})&\cellcolor{gray!25}\textit{LogLoss (9)}&\cellcolor{gray!25}\textit{AUC (10)}&\cellcolor{gray!25}\textit{LogLoss (10)}\\ \hline
\cellcolor{red!5}\textit{LR}          & 0.747349  & 0.379716      & 0.739194   & 0.401167       \\ \hline
\cellcolor{red!5}\textit{GNN}         & 0.753254  & 0.377235      & 0.739651   & 0.400147       \\ \hline
\cellcolor{green!5}$\text{\textit{Esb}}_{\text{\textit{Magnitude}}}$   & 0.756016  & 0.375891      & 0.745535   & 0.414997       \\ \hline
\cellcolor{green!5}$\text{\textit{Esb}}_{\text{\textit{Consistency}}}$ & 0.752616  & 0.383943      & 0.744574   & 0.410970       \\ \hline
\cellcolor{green!5}$\text{\textit{Esb}}_{\text{\textit{Contrast}}}$    & 0.756796  & 0.375351      & \textbf{0.746868}   &\textbf{0.397326}       \\ \hline
\cellcolor{green!5}$\text{\textit{Esb}}_{\text{\textit{Correlation}}}$ & 0.756187  & 0.375575      & 0.745156   & 0.398449       \\ \hline
\cellcolor{green!5}$\text{\textit{Esb}}_{\text{\textit{Overall}}}$     & \textbf{0.757125}  &\textbf{0.375096}      & 0.745980   & 0.397525       \\ \hline
\end{tabular}
\label{tbl:fwls}
\end{table}

Tab.~\ref{tbl:fwls} shows the performance of the original and ensembled models. We have three findings. First, all ensembled models (except $\text{\textit{Esb}}_{\text{\textit{Consistency}}}$ for Day 9) are better than the original \textit{LR} and \textit{GNN}, reflecting the efficacy of FWLS. 
Second, for both Day 9 and Day 10 data, $\text{\textit{Esb}}_{\text{\textit{Overall}}}$ is better than $\text{\textit{Esb}}_{\text{\textit{Magnitude}}}$, indicating our \textit{Overall} metric that profiling features' importance from multiple perspectives is better than the traditional \textit{Magnitude} metric. 
Third, for Day 10 data, $\text{\textit{Esb}}_{\text{\textit{Contrast}}}$ generates the best performance, indicating that the weights in our \textit{Overall} metric (Eq.~\ref{eq:overall}) may need to be adjusted for different datasets.

\subsection{Ablation Study on Meta-Features}
\label{sec:ablation}
Meta-features are dominating factors controlling the quality of the discriminator (\textit{Step 5} of LFD). Therefore, using the discriminator from Sec.~\ref{sec:ensemble}, we conduct ablation studies to investigate how the \textit{number} and \textit{quality} of meta-features affect the discriminator's performance.

\setlength{\belowcaptionskip}{-5pt}
\begin{figure}[b]
 \centering 
 \includegraphics[width=\columnwidth]{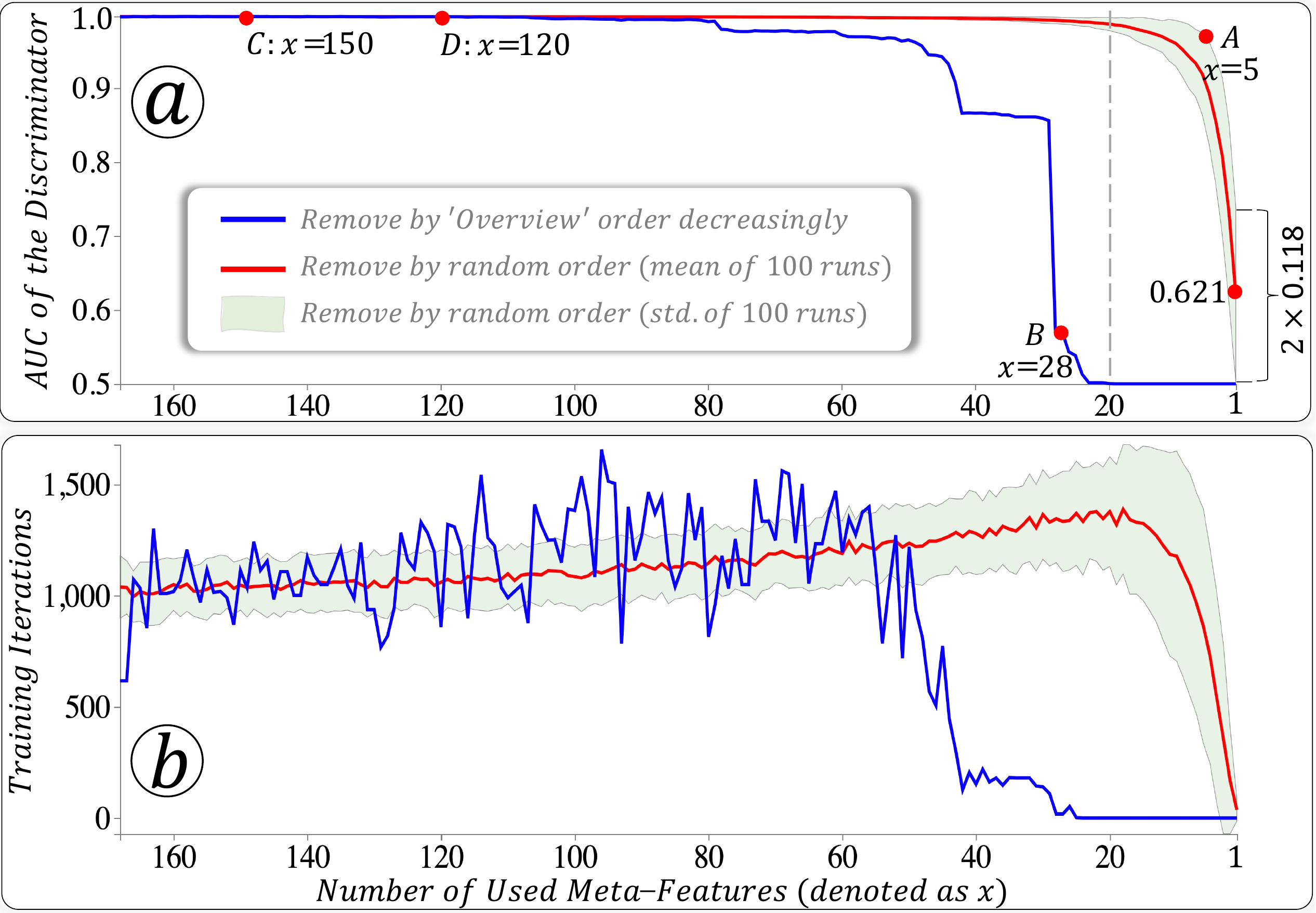}
  \vspace{-0.2in}
 \caption{Ablation study to investigate how the \textit{Discriminator}'s performance is affected by the number and quality of of meta-features.}
 \label{fig:ablation}
\end{figure}
\setlength{\belowcaptionskip}{0pt}

To reveal how many meta-features are sufficient to train a good discriminator, we remove the 168 meta-features one-by-one, retrain the discriminator from scratch after each removal, and measure its performance by AUC. What order to use when removing the meta-features is an important question. To simulate the possible choices of meta-features with different qualities, we remove them in a random order but repeat the entire ablation process 100 times. The red curve in Fig.~\ref{fig:ablation}a shows how the average AUC (averaged over the 100 runs) changes along with the decreasing number of meta-features. The green band surrounding the red curve shows the standard deviation of the 100 runs. For example, the right-most point on the red curve shows the average AUC of the discriminator (when being trained with one meta-feature only) is 0.621, and the standard deviation is 0.118. The upper/lower bound of the green band reflects the case of using high/low-quality meta-features. The blue curve in Fig.~\ref{fig:ablation}a shows the AUC of the discriminator when removing meta-features following our \textit{Overall} order. From the figure, we can draw the following conclusions.
\begin{compactitem}
\item Our \textit{Overall} metric successfully prioritizes important meta-features, as removing features following it makes the AUC drop much earlier and more significantly.
\item Good meta-features, even only 5 of them (Fig.~\ref{fig:ablation}a, point $A$), can train a good discriminator to differentiate the two compared models. Low-quality meta-features, as many as 28 (Fig.~\ref{fig:ablation}a, point $B$), are less effective to train the discriminator. So, meta-features' quality matters.
\item Some meta-features or their combinations may have similar effects. For example, combing  meta-feature $\beta$ and $\gamma$ may produce the equivalent effect of feature $\alpha$. Consequently, removing $\alpha$ will not significantly affect the discriminator's quality. This could be the reason for the very similar AUC at point $C$ and $D$ in Fig.~\ref{fig:ablation}a. More analysis regarding this can be found in our Appendix.
\end{compactitem}

When training the discriminator (i.e., an XGBoost), we set the training iterations to be very large but terminate the training if the AUC has no improvement in 10 consecutive iterations. This mechanism makes sure the model is sufficiently trained. Following the encoding in Fig.~\ref{fig:ablation}a, Fig.~\ref{fig:ablation}b shows the number of training iterations. It can be seen that the number of meta-features does not obviously affect the number of training iterations if the discriminator can be successfully trained (i.e., mostly ${\sim}1000$ iterations). However, when the discriminator cannot be successfully trained (due to the insufficient number or low quality of meta-features), the number of training iterations drops significantly.

From this study, we notice that it is important to reveal the quality of the discriminator. For different cases, however, it is hard to know how many meta-features are sufficient and what qualities the meta-features have.
Therefore, we reflect the AUC of the trained discriminators in the system (i.e., the numbers in Fig.~\ref{fig:system}-b8 show the AUC of the TP and FP discriminators). This quality indicator helps users get to know if their meta-features are sufficient or good enough before they can interpret the SHAP visualizations.

\textit{\textbf{How hard to train the discriminators?}} Theoretically, the discriminators' job is much easier than model \textit{A} and \textit{B}, as both can clearly separate the $\text{\textit{A}}^\text{+}\text{\textit{B}}^\text{-}$ and $\text{\textit{A}}^\text{-}\text{\textit{B}}^\text{+}$ instances using their score cutoff, e.g., for model $A$, all $\text{\textit{A}}^\text{+}\text{\textit{B}}^\text{-}$ instances are from $\text{\textit{A}}^\text{+}$ (above the cutoff) and all $\text{\textit{A}}^\text{-}\text{\textit{B}}^\text{+}$ instances are from $\text{\textit{A}}^\text{-}$ (below the cutoff). Practically, this can also be verified from the much better AUCs of the discriminators (${>}0.99$ in Fig.~\ref{fig:system}-b8) than the two compared models (${<}0.76$ in Tab.~\ref{tbl:fwls}).

\subsection{Domain Experts' Feedback}
\label{sec:feedback}
We conduct multiple case studies (Sec.~\ref{sec:casestudy}) with the eight ML experts ($E_1{\sim}E_8$) introduced in Sec.~\ref{sec:requirement}. The four merchant category verification models in Sec.~\ref{sec:vda} are proposed by the experts and they have sufficient knowledge on their differences. As a sanity check, the insights derived from LFD match well with the experts' expectations, e.g., the \textit{RNN} is better than the \textit{Tree} in capturing temporal behaviors, the affinity information helps the \textit{GNN} outperform the \textit{RNN}. The comparison of CTR models also makes sense to the experts and the proposed meta-features (e.g., \textit{\textbf{n\_active\_hours}\_*}) help to reveal the \textit{RNN}'s deficiency on short sequences. We conclude the studies with open-ended interviews and think-aloud discussions, and summarize the experts' feedback from the following perspectives.

\textbf{\textit{LFD Idea and Its Understandability.}} In general, all experts felt positive about the idea to learn from two models' disagreement. 
$E_1$ liked the idea a lot and considered LFD as an ``\textit{offline adversarial learning}'' (in analogy to the online adversarial learning of GAN~\cite{gan2014goodfellow}), where the two compared models are used to explicitly identify where to learn and our discriminator is similar to the discriminator of GAN. He also identified the advantage of LFD in ``\textit{smartly avoiding the data imbalance issue}''. 
$E_6$ stated that LFD is ``\textit{similar to existing Explainable AI works}'' that ``\textit{train a surrogate model by distilling knowledge from original models}''. He considered our discriminator as the surrogate which distills disagreement knowledge from the comparison. 
$E_7$ commented that the ``\textit{LFD pipeline is smooth without any gap}''. $E_8$ echoed that the novel part of LFD is to use ``\textit{supervised learning}'' to relate users' interested model-behaviors (externalized by meta-features) with the models' prediction discrepancy.
Regarding the difficulty level of LFD, most experts agreed that the overall idea is straightforward and Fig.~\ref{fig:pipeline} is a good illustration. $E_6$ and $E_8$ expected the logic of \textit{Step 4} to be difficult for novices to comprehend, and commented that users do need some deeper thoughts to understand the complicated disagreement scenarios in real-world applications.

\textit{\textbf{Meta-Features and Feature-Ordering.}} 
Some experts understood the concept of meta-features very easily. For example, $E_3$ worked on FWLS and immediately accepted this concept, as it was also used in FWLS. 
The concept was new to some other experts, e.g., $E_6$. After understanding it, $E_6$ differentiated meta-features from training features based on their purpose, i.e., ``\textit{training features are to improve models' performance}'' whereas ``\textit{meta-features are to improve models' interpretability}''. 
About how hard it is to generate meta-features, $E_7{\sim}E_8$ commented that they often have thousands of ideas to generate features, as long as they are familiar with the data. However, they were concerned about whether the quality of their meta-features would be good enough.
We comforted them that they can blindly throw all their meta-features into LFD, use the feature-ordering metrics to sort them, and compare models based on the important meta-features only. $E_7$ advocated this idea and considered it as a good feature filtering mechanism. 
About our feature-ordering metrics, all experts supported our consideration to more comprehensively rank features. $E_2$ has decades of ML experience. He commented that ``\textit{the feature order is extremely important}'' in his model building process and appreciated the new metrics. $E_7$ felt the \textit{Contrast} order should be the best. We presented her a meta-feature with large \textit{Contrast} but small \textit{Magnitude} to show how features' magnitude could also limit their contribution. 
$E_8$ worked on similar feature ordering problems and felt our metrics ``\textit{very inspiring}''.

\textit{\textbf{Visual Interface and SHAP Interpretation.}}
For the \textit{Disagreement Distribution View}, most experts could easily get our design philosophy after watching the associated video illustrations. However, some experts tended to ignore the disagreement distributions, but chose the threshold (for \textit{Step 2} of LFD) based on their experience, to make the analysis more realistic. For the \textit{Feature View}, the interactions were mostly ordering the meta-features and clicking individual ones to check their details in the bubble-plot. These interactions and the ``\textit{overview+details}'' exploration were intuitive to the experts. $E_6$ focused on the color legend (in Fig.~\ref{fig:system}-b3) during explorations and liked the flexible color mapping function (i.e., by dragging the control points). He found this widget very useful, especially when the contrast between the left and right half of the bubble-plot is not obvious. He also recommended algorithms to automatically find the best color mapping. $E_7$ felt the numerical values in Fig.~\ref{fig:system}-b6 are sufficient to show the orders and suggested removing the blue bars (with orange triangles) to save space. 
Although our eight users are ML experts, not all of them are experts in SHAP. However, as LFD uses SHAP as a ``black-box", all experts felt comfortable when interpreting the outcomes. Also, we would like to admit the SHAP dependency of our framework and the corresponding limitations (see Sec.~\ref{sec:discussion}).

\textit{\textbf{Limitations.}} First, there was a discussion on the possibility of extending LFD to compare more than two classifiers. The straightforward extension is to train a multi-class discriminator at \textit{Step 5} of LFD. $E_6$ pointed out the complexity would likely come from the disagreement matrix, where the number of cells is roughly $2^n$ ($n$ is the number of compared models), which may make the logic complicated. Another limitation identified by $E_2$ and $E_6$ is that the two compared models must have disagreements before LFD can be executed. If $\text{\textit{A}}^\text{+}{\subset}\text{\textit{B}}^\text{+}$ or $\text{\textit{B}}^\text{+}{\subset}\text{\textit{A}}^\text{+}$ (at \textit{Step 2}), LFD can learn nothing. 
Third, $E_7$ and $E_8$ discussed how different types of discriminators would affect the interpretability of LFD. We currently used XGBoost but other SHAP-friendly models could also be used, and we agreed it is a good future direction to compare the interpretation outcomes from different discriminators. Lastly, $E_8$ also mentioned the possibility of contradictory feature conditions.
He continued our spam filtering example and explained that the coming emails may have large $n\_url$ and large $n\_cap$. From LFD, we may know that model $\text{\textit{A}}$ outperforms $\text{\textit{B}}$ when $n\_url$ is large, whereas $\text{\textit{B}}$ outperforms $\text{\textit{A}}$ when $n\_cap$ is large. In this case, it is hard to choose between $\text{\textit{A}}$ and $\text{\textit{B}}$ as LFD would recommend both. We did not see such cases in our current experiments but were glad to notice such possible scenarios. A quick remedy is to vote models based on the meta-features. We would like to conduct more studies in this direction in the future.

\textit{\textbf{Suggested Improvements.}} We asked the experts to provide suggestions or desired functions to improve LFD at the end. $E_1{\sim}E_2$ encouraged us to integrate more feature interpretation methods to make LFD more general. $E_3{\sim}E_4$ wanted to see the precision of the two compared models at different thresholds in the \textit{Disagreement Distribution View} (Fig.~\ref{fig:system}-a4). $E_5$ asked us to extend LFD by enabling ``\textit{the negative side comparisons}'' (i.e., comparing true-negative and false-negative instances by sorting instances increasingly at \textit{Step 2}). $E_6$ recommended us weighting the disagreed instances based on their disagreement-level before learning from them. This is similar to LIME on weighting instances based on their locality to the interpreted instance. $E_7{\sim}E_8$ preferred to employ learning-to-rank algorithms to generate the best weights in the \textit{Overall} metric. These suggestions provide promising directions for us to improve LFD.

\section{Discussion, Limitations, and Future Work}
\label{sec:discussion}

\textit{\textbf{The $\text{\textit{A}}^\text{-}\text{\textit{B}}^\text{-}$ cell.}} We focused on positive predictions (i.e., TP and FP instances) of the two compared classifiers, and thus, the instances are at least captured by one model. As a result, we do not have the $\text{\textit{A}}^\text{-}\text{\textit{B}}^\text{-}$ cell in LFD. However, as suggested by $E_5$ (Sec.~\ref{sec:feedback}), one can also compare two classifiers from the negative predictions, i.e., sorting the scores increasingly at \textit{Step} 2 of LFD. This is less common and out of the scope of this paper. Also, we want to emphasize that the $\text{\textit{A}}^\text{-}\text{\textit{B}}^\text{-}$ (as well as the $\text{\textit{A}}^\text{+}\text{\textit{B}}^\text{+}$) cell is of less interest for the purpose of comparison, as it is where the two classifiers agree on.

\textit{\textbf{Dependency on SHAP.}} LFD depends on SHAP, so does the interpretation accuracy. Consequently, it has an inherent limitation on cases where SHAP cannot provide accurate interpretations. For this limitation, we want to emphasize two points. First, SHAP is widely applicable to most ML models and comes with solid theoretical supports~\cite{lundberg2018consistent,NIPS2017_7062}. So, we do not expect considerably inaccurate interpretations to occur very often. Second, as the six steps of LFD have been very well modularized, we can easily replace SHAP with other interpretation methods at \textit{Step 6}.

There are three \textit{\textbf{scalability}} concerns of LFD. First, LFD is limited to the comparison between two binary classifiers only due to our focused problem (i.e., ``swap analysis''). We discussed the possibility of extending LFD to compare more classifiers (Sec.~\ref{sec:feedback}) and considered it as an important future work. For multi-class classifiers, one can still use LFD to compare their difference on a single class at a time. Second, our system currently supports hundreds of meta-features. However, for cases with thousands of meta-features, our visualization (the overview part, Fig.~\ref{fig:system}-b5) may not scale well. Fortunately, using the proposed new metrics, we can eliminate the less important meta-features from visualization to reduce the cost. Lastly, the SHAP computations may also be a bottleneck when the data is large. However, SHAP can be computed offline and one may consider other more efficient model interpretation methods for a replacement.

We believe \textit{\textbf{LFD is useful}} from at least two perspectives. 
First, it provides feature-level interpretations when comparing two classifiers (i.e., \textit{comparative interpretations}), providing unique insights into model selections. 
We want to emphasize that LFD is not designed to completely replace or compete against any existing model comparison solutions. Existing solutions, e.g., comparing models with aggregated metrics, still have their advantages (e.g., simpler and more generalizable). We expect LFD to be used together with them and reveal models' discrepancy from a different angle.
Second, LFD provides more effective metrics in prioritizing meta-features, leading to better model ensembling. As explained, the importance of features is often measured by their contribution \textit{Magnitude}, which depicts features' importance from one perspective only. Our \textit{Overall} metric profiles features from multiple perspectives and could more comprehensively reflect features' importance.

\section{Conclusion}

This work introduces LFD, a model comparison and visualization framework. LFD compares two classification models by identifying data instances with disagreed predictions, and learning from their disagreement using a set of user-proposed meta-features. Based on the SHAP interpretation of the learning outcomes, we can interpret the two models' behavior discrepancy on different feature conditions. Also, we propose multiple metrics to prioritize the meta-features and use model ensembling to evaluate the metrics' quality. Through concrete cases conducted with ML experts on real-world problems, we validate the efficacy of LFD.

\ifCLASSOPTIONcaptionsoff
  \newpage
\fi



\bibliographystyle{IEEEtran}
\bibliography{IEEEabrv,./tvcg21}
%
%
%

%

\vspace{-0.3in}
\begin{IEEEbiography}[{\includegraphics[width=1in,height=1.25in,clip,keepaspectratio]{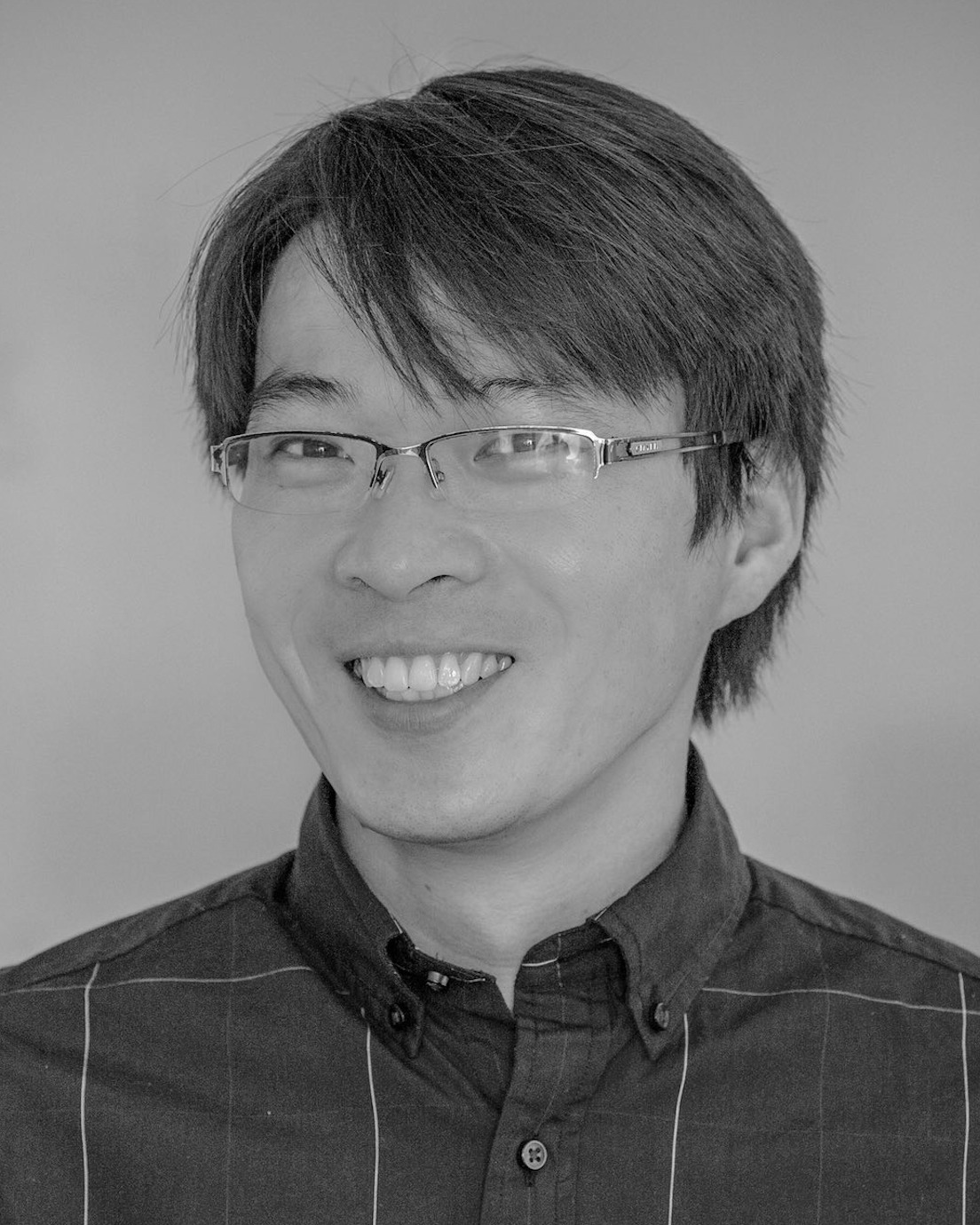}}]{Junpeng Wang}
is a staff research scientist at Visa Research. He received his B.E. degree in software engineering from Nankai University in 2011, M.S. degree in computer science from Virginia Tech in 2015, and Ph.D. degree in computer science from the Ohio State University in 2019. His research interests are broadly in visualization, visual analytics, and explainable AI.
\end{IEEEbiography}

\vspace{-0.3in}
\begin{IEEEbiography}[{\includegraphics[width=1in,height=1.25in,clip,keepaspectratio]{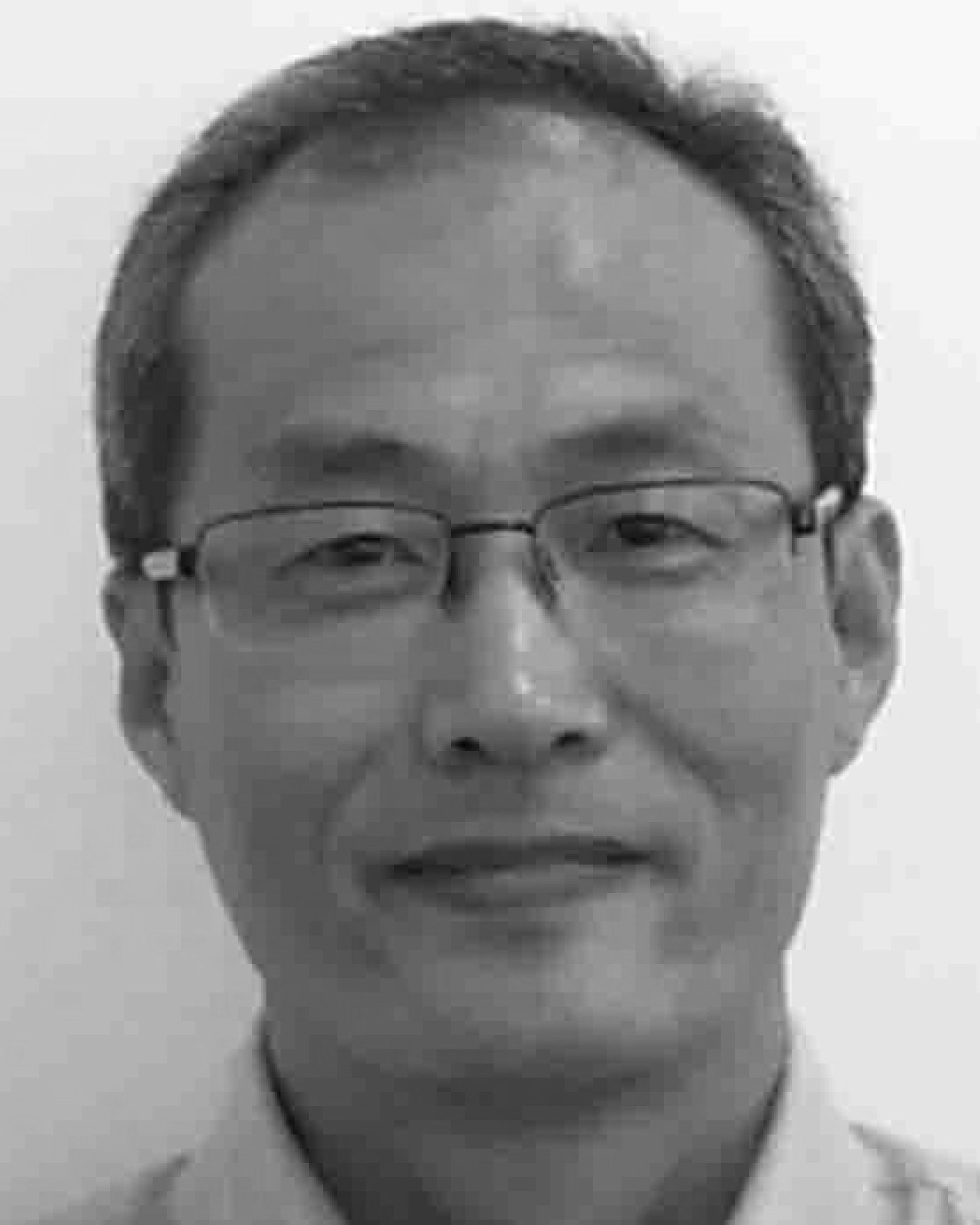}}]{Liang Wang}
is a principal research scientist in the Data Analytics team at Visa Research.  His research interests are in data mining, machine learning, and fraud analytics. Liang received his Ph.D. degree in Computer Science from Facult\'e Polytechnique de Mons, Mons, Belgium with highest honors, and his BS degree in Electrical Engineering \& Automation and his MS degree in Systems Engineering, both from Tianjin University. Prior to joining Visa, He was a senior principal data scientist at Yahoo!, responsible for building traffic protection solutions for Yahoo!’s advertising platforms. Before Yahoo!, he was a distinguished scientist at eBay/PayPal, leading projects on risk detection for PayPal’s core payment system. Liang also worked at FICO as a senior scientist, focusing on bankcard fraud detection. He is the inventor of over 20 patents and has published over 30 papers in international journals and conferences.
\end{IEEEbiography}

\vspace{-0.3in}
\begin{IEEEbiography}[{\includegraphics[width=1in,height=1.25in,clip,keepaspectratio]{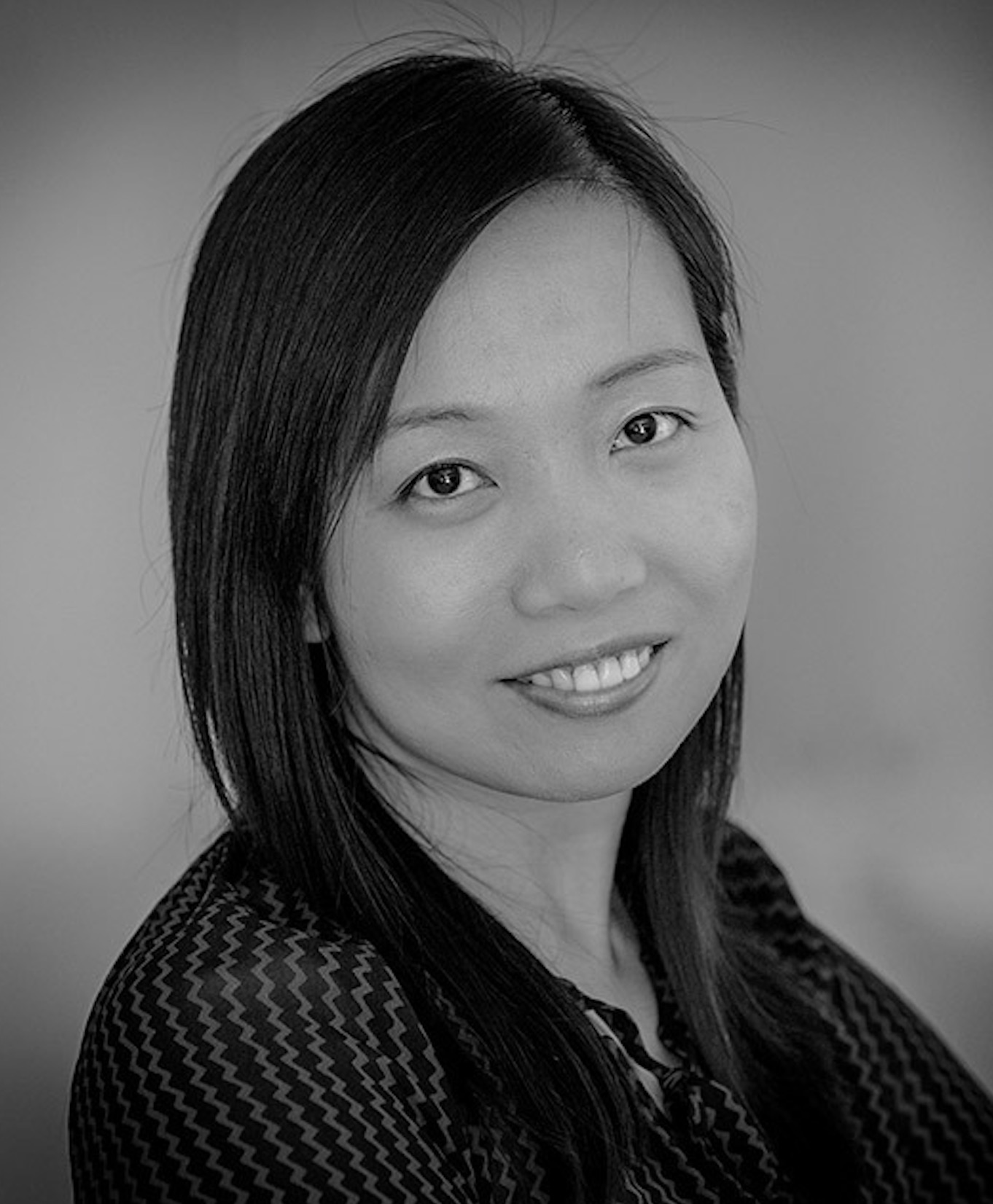}}]{Yan Zheng} is currently a Senior Staff Research Scientist at Visa Research. Yan received her Ph.D. in Computer Science from University of Utah in 2017. She has published papers in top venues, including KDD, SIGMOD, ICDM and others. Her research interests are in data mining, machine learning, and representation learning.
\end{IEEEbiography}

\vspace{-0.3in}
\begin{IEEEbiography}[{\includegraphics[width=1in,height=1.25in,clip,keepaspectratio]{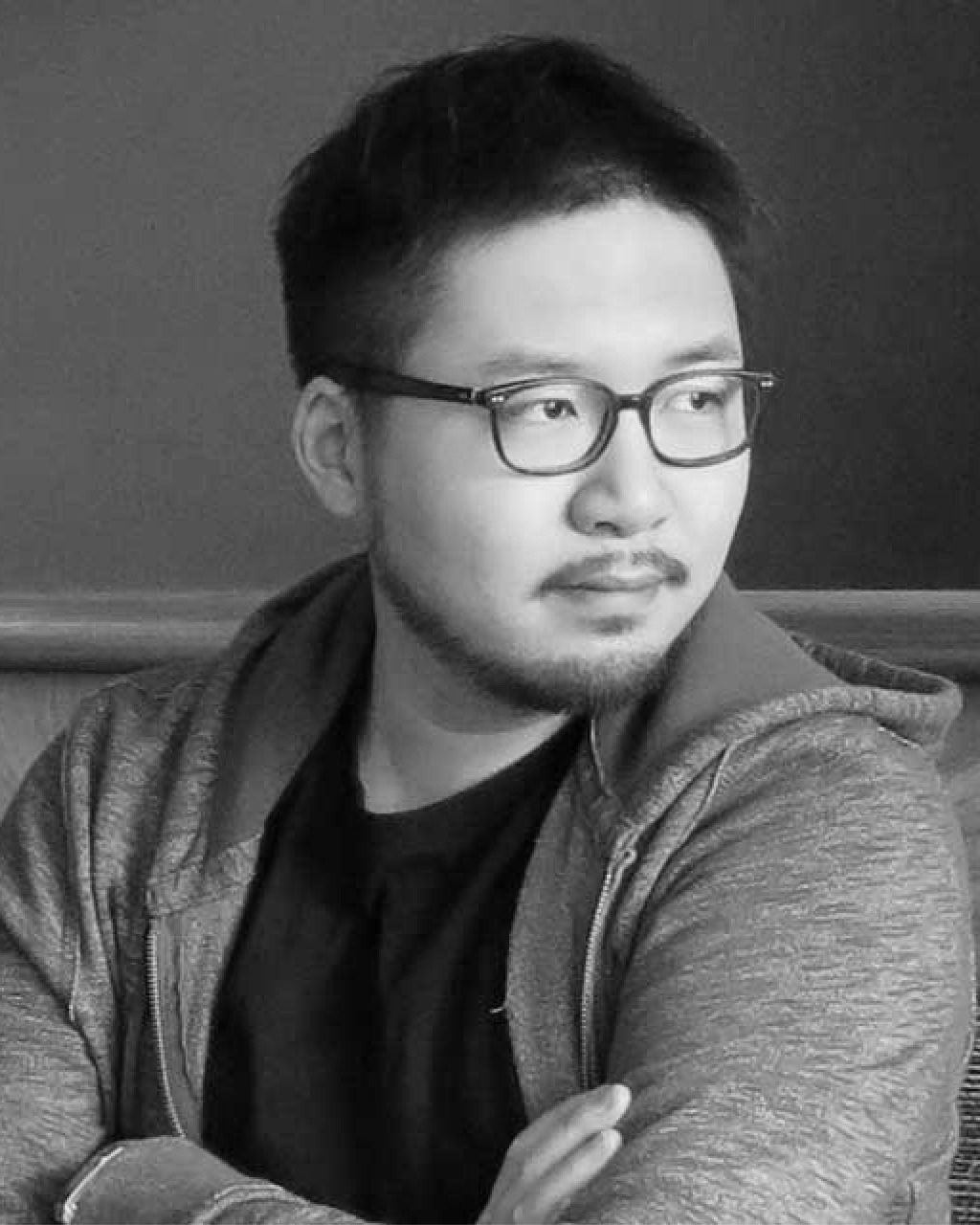}}]{Chin-Chia Michael Yeh}
is currently a Staff Research Scientist at Visa Research. Michael received his Ph.D. in Computer Science from University of California, Riverside. His Ph.D. thesis “Toward a Near Universal Time Series Data Mining Tool: Introducing the Matrix Profile,” received Doctoral Dissertation Award Honorable Mention at KDD 2019. He has published papers in top venues, including KDD, VLDB, ICDM and others. His research interests are in data mining, machine learning, and time series analysis.
\end{IEEEbiography}

\vspace{-0.3in}
\begin{IEEEbiography}[{\includegraphics[width=1in,height=1.25in,clip,keepaspectratio]{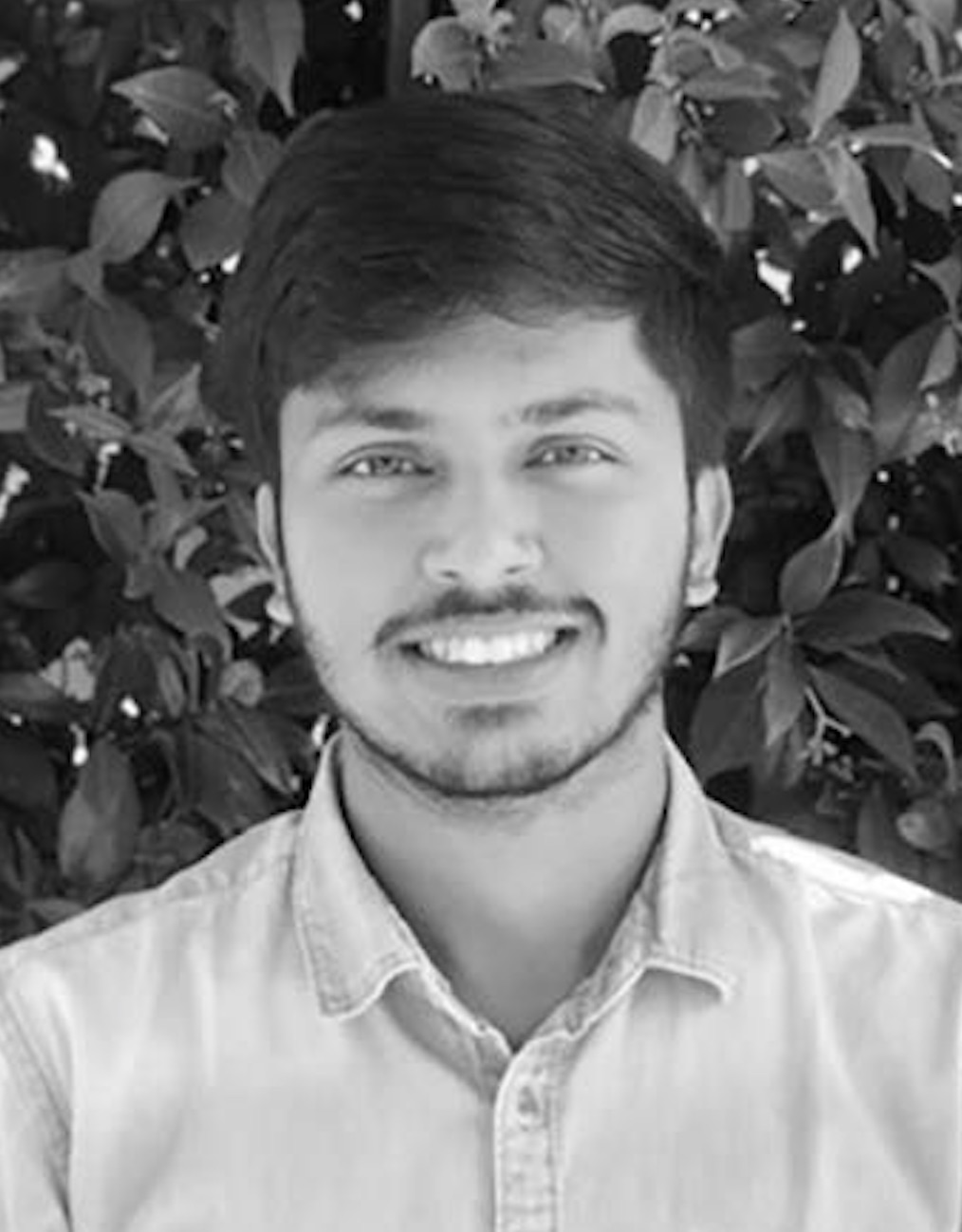}}]{Shubham Jain} is a Staff Software Engineer at Visa Research. He received his M.S. in Computer Science from University of Illinois at Urbana-Champaign in 2019, and received his B.T. in Computer Science and Engineering from Indian Institute of Technology Kanpur in 2017. His master’s thesis focused on using landmarks for enhancing cloth retrieval. Prior to joining Visa, Shubham was a Software Engineer Intern at NVIDIA, working on deep learning research. Before working at NVIDIA, he was a Research Intern at Adobe and a Visiting Student Researcher at Montreal Institute of Learning Algorithms (MILA) under Prof. Yoshua Bengio. His interests broadly lies in Computer Vision, Recurrent Neural Networks and Machine Learning.
\end{IEEEbiography}

\vspace{-0.3in}
\begin{IEEEbiography}[{\includegraphics[width=1in,height=1.25in,clip,keepaspectratio]{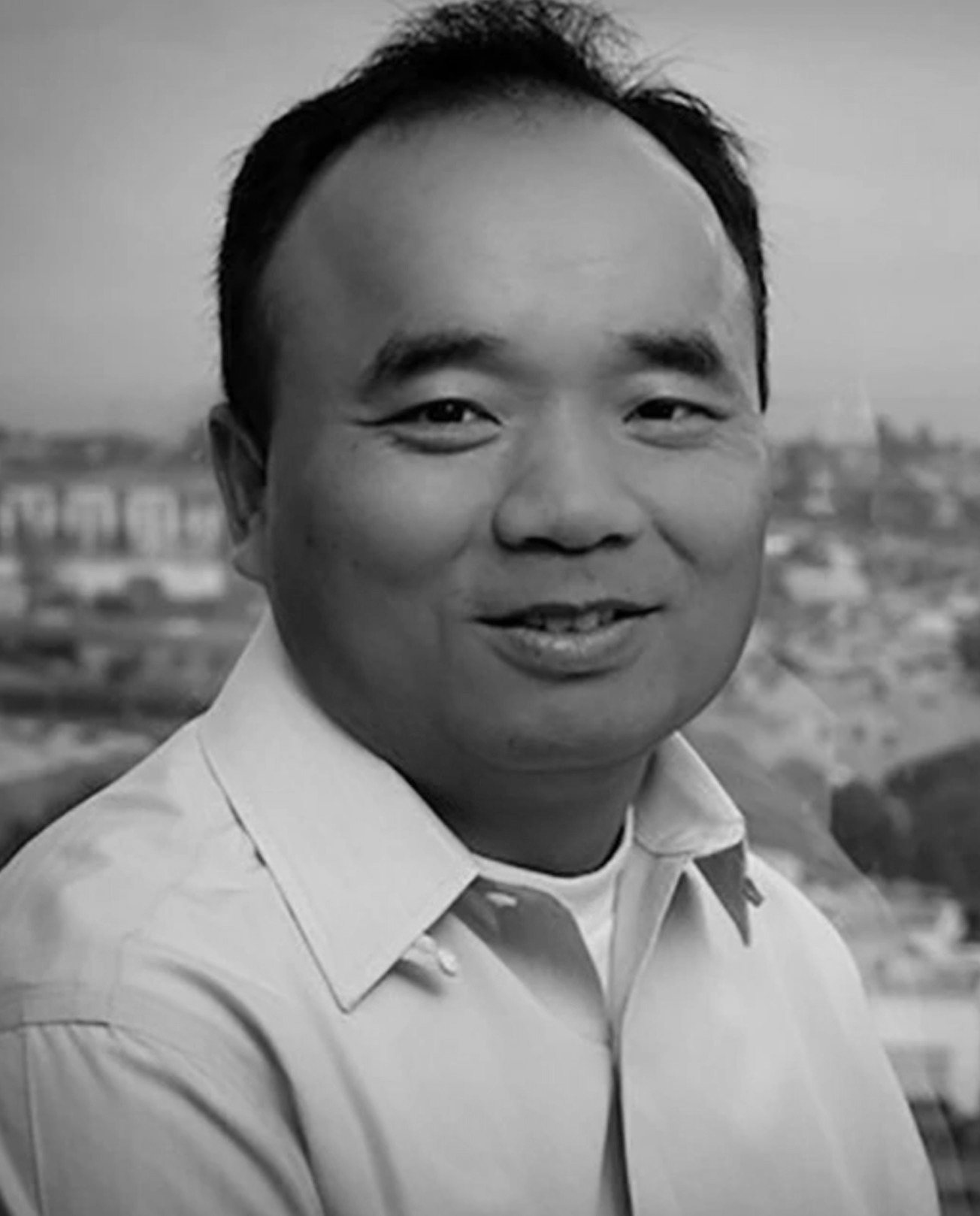}}]{Wei Zhang}
is a principal research scientist and research manager at Visa Research and interested in big data modeling and advanced machine learning technologies for payment industry. Prior to joining Visa Research, Wei worked as a Research Scientist in Facebook, R\&D manager in Nuance Communications and also worked in IBM research over 10 years. Wei received his Bachelor and Master degrees from Department of Computer Science, Tsinghua University.
\end{IEEEbiography}




\end{document}